\documentclass[sigconf]{acmart}
\AtBeginDocument{%
  }

\setcopyright{acmlicensed}
\copyrightyear{2026}
\acmYear{2026}
\acmDOI{XXXXXXX.XXXXXXX}
\acmConference[MM '26]{Make sure to enter the correct
  conference title from your rights confirmation email}{November 10--14,
  2026}{Rio de Janeiro, Brazil}
\acmISBN{978-1-4503-XXXX-X/2018/06}

\acmSubmissionID{8103}

\usepackage{listings}
\usepackage{xcolor}
\usepackage{multirow}
\usepackage{makecell}

\definecolor{jsonkey}{rgb}{0.0,0.3,0.7}
\definecolor{jsonstring}{rgb}{0.6,0.1,0.1}
\definecolor{jsonnumber}{rgb}{0.1,0.5,0.1}

\lstdefinelanguage{json}{
    basicstyle=\ttfamily\footnotesize,
    numbers=left,
    numberstyle=\tiny\color{gray},
    stepnumber=1,
    numbersep=8pt,
    showstringspaces=false,
    breaklines=true,
    frame=single,
    rulecolor=\color{black},
    backgroundcolor=\color{gray!5},
    literate=
     *{:}{{{\color{black}:}}}{1}
      {,}{{{\color{black},}}}{1}
      {"}{{{\color{black}"}}}{1}
      {true}{{{\color{jsonnumber}true}}}{4}
      {false}{{{\color{jsonnumber}false}}}{5}
      {null}{{{\color{jsonnumber}null}}}{4},
    stringstyle=\color{jsonstring},
    keywordstyle=\color{jsonkey}
}

\lstdefinestyle{idslstyle}{
  basicstyle=\fontsize{7.5pt}{8pt}\ttfamily,  % ← 强制自定义字号
  numbers=none
}

\usepackage{algorithm}
\usepackage{algpseudocode}

\usepackage{booktabs}
\usepackage{pifont}     % for \ding
\usepackage{graphicx}   % for \resizebox
\newcommand{\cmark}{\ding{51}} % ✓
\newcommand{\xmark}{\ding{55}} % ✗
\begin{document}

%%
%% The "title" command has an optional parameter,
%% allowing the author to define a "short title" to be used in page headers.
\title{RoomPilot: Controllable Indoor Scene Synthesis via Multimodal Semantic Parsing}

%%
%% The "author" command and its associated commands are used to define
%% the authors and their affiliations.
%% Of note is the shared affiliation of the first two authors, and the
%% "authornote" and "authornotemark" commands
%% used to denote shared contribution to the research.
\author{Wentang Chen}
\authornote{Both authors contributed equally to this research.}
\email{b241000657@hnu.edu.cn}
\affiliation{%
  \institution{School of Information Science and Engineering, Hunan University}
  \city{Changsha}
  \state{Hunan}
  \country{China}
}

\author{Shougao Zhang}
\authornotemark[1]
\email{zhangshougao@hnu.edu.cn}
\affiliation{%
  \institution{School of Information Science and Engineering, Hunan University}
  \city{Changsha}
  \state{Hunan}
  \country{China}
}

\author{Yiman Zhang}
\email{zhangyiman@hnu.edu.cn}
\affiliation{%
  \institution{School of Information Science and Engineering, Hunan University}
  \city{Changsha}
  \state{Hunan}
  \country{China}
}

\author{Tianhao Zhou}
\email{butter@hnu.edu.cn}
\affiliation{%
  \institution{School of Information Science and Engineering, Hunan University}
  \city{Changsha}
  \state{Hunan}
  \country{China}
}

\author{Ruihui Li}
\authornote{Corresponding author.}
\email{liruihui@hnu.edu.cn}
\affiliation{%
  \institution{School of Information Science and Engineering, Hunan University}
  \city{Changsha}
  \state{Hunan}
  \country{China}
}

%%
%% By default, the full list of authors will be used in the page
%% headers. Often, this list is too long, and will overlap
%% other information printed in the page headers. This command allows
%% the author to define a more concise list
%% of authors' names for this purpose.
\renewcommand{\shortauthors}{Trovato et al.}

%%
%% The abstract is a short summary of the work to be presented in the
%% article.
\begin{abstract}
% Generating controllable and interactive indoor scenes is fundamental to applications in game development, architectural visualization, and embodied AI training. Yet existing approaches either handle a narrow range of input modalities or rely on stochastic processes that hinder controllability. To overcome these limitations, we introduce RoomPilot, a unified framework that parses diverse multi-modal inputs— textual descriptions or CAD floor plans—into an Indoor Domain-Specific Language (IDSL) for indoor structured scene generation. The key insight is that a well-designed IDSL can act as a shared semantic representation, enabling coherent, high-quality scene synthesis from any single modality while maintaining interaction semantics. In contrast to conventional procedural methods that produce visually plausible but functionally inert layouts, RoomPilot leverages a curated dataset of interaction-annotated assets to synthesize environments exhibiting realistic object behaviors. Extensive experiments further validate its strong multi-modal understanding, fine-grained controllability in scene generation, and superior physical consistency and visual fidelity, marking a significant step toward general-purpose controllable 3D indoor scene generation.
Generating controllable indoor scenes is fundamental to applications in game development, architectural visualization, and embodied AI. However, existing approaches either support a limited input modalities or rely on implicit generation processes that hinder precise control over scene structure and semantics. To address these limitations, we introduce RoomPilot, a unified framework for controllable indoor scene synthesis from multi-modal inputs, including textual descriptions and CAD floor plans. RoomPilot maps heterogeneous inputs into an Indoor Domain-Specific Language (IDSL), which serves as a structured and interpretable semantic representation for describing indoor scenes. Built upon IDSL, RoomPilot presents a hierarchical synthesis pipeline that progressively organizes scenes at the building, room, and object levels, promoting structural coherence and functional consistency across multi-room layouts. Moreover, RoomPilot constructs a curated asset dataset with rich semantic annotations to support high-quality scene synthesis, improving visual realism and appearance consistency. Extensive experiments demonstrate effective multi-modal understanding, fine-grained controllability in scene generation, and improved physical consistency and visual fidelity, marking a significant step toward controllable 3D indoor scene synthesis. Code and model will be available.
\end{abstract}

%%
%% The code below is generated by the tool at http://dl.acm.org/ccs.cfm.
%% Please copy and paste the code instead of the example below.
%%



%%
%% Keywords. The author(s) should pick words that accurately describe
%% the work being presented. Separate the keywords with commas.
%% A "teaser" image appears between the author and affiliation
%% information and the body of the document, and typically spans the
%% page.
\begin{teaserfigure}
  \includegraphics[width=\textwidth]{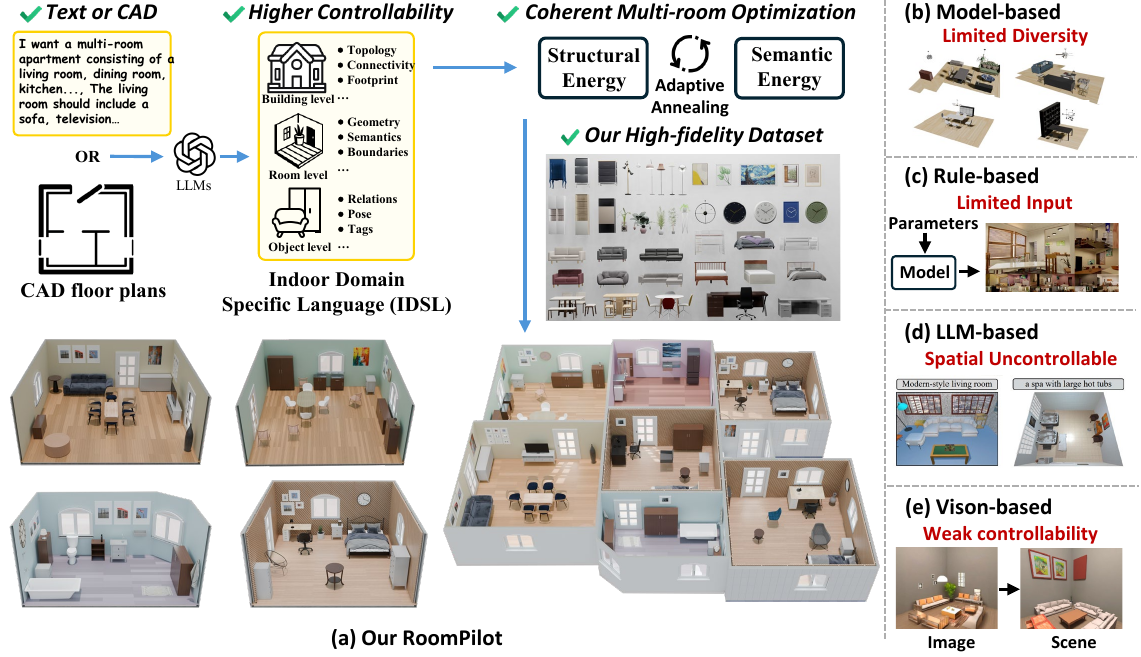}
  \caption{
  % Overview of RoomPilot.
  (a) Our RoomPilot converts text or CAD into an IDSL and performs coherent multi-room optimization to generate controllable indoor scenes, while (b) model-based methods show limited diversity, (c) rule-based methods support restricted inputs, (d) LLM-based methods lack spatial controllability, and (e) vision-based methods provide weak controllability.}
  \label{fig:teaser}
\end{teaserfigure}

\received{20 February 2007}
\received[revised]{12 March 2009}
\received[accepted]{5 June 2009}

%%
%% This command processes the author and affiliation and title
%% information and builds the first part of the formatted document.
\maketitle

\section{Introduction}
\label{sec:intro}

\begin{table*}[t]
\centering
\caption{\textbf{Comparison of different scene synthesis methods.}
We compare methods in terms of multi-room support, controllability, CAD source, detailed scene descriptions, spatial relations, physical plausibility, and supported input modalities.
\textit{Controllability} refers to whether a method allows users to specify or constrain scene layouts beyond implicit generation,
while \textit{Spatial Relations} indicate explicit specification of relative or functional spatial relationships (e.g., adjacency, facing, or on-top).
% RoomPilot enables relation-aware and physically plausible multi-room scene synthesis.
}
\resizebox{\textwidth}{!}{
\begin{tabular}{lccccccc}
\toprule
\textbf{Method} 
& \textbf{Multi-room} 
& \textbf{Controllability} 
& \textbf{CAD Source} 
& \textbf{Detailed Descriptions} 
& \textbf{Spatial Relations}
& \textbf{Physical Plausibility} 
& \textbf{Input Modalities} \\
\midrule
ATISS~\cite{paschalidou2021atiss}                 
& \xmark & \xmark & 3D-FUTURE \cite{fu20213d} & \xmark & \xmark & \xmark & Scenes \\

DiffuScene~\cite{tang2023diffuscene}   
& \xmark & \cmark & 3D-FUTURE \cite{fu20213d} & \xmark & \xmark & \xmark & Text (+ partial scene) \\

PhyScene~\cite{yang2024physcene}           
& \cmark & \cmark & 3D-FUTURE \cite{fu20213d} & \xmark & \xmark & \cmark & Text / task priors \\

Infinigen Indoors~\cite{infinigen2024indoors}  
& \cmark & \xmark & Generated & \xmark & \xmark & \cmark & Procedural spec \\

ProcTHOR~\cite{procthor}           
& \cmark & \xmark & RoboTHOR \cite{deitke2020robothor} & \xmark & \xmark & \cmark & API spec \\

MetaScenes~\cite{yu2025metascenes}       
& \cmark & \xmark & Mixed & \xmark & \xmark & \cmark & Scan + CAD assets \\

ACDC~\cite{dai2024automated}                   
& \cmark & \xmark & Behavior & \xmark & \xmark & \cmark & Image \\

Architect~\cite{wangarchitect}         
& \cmark & \cmark & Mixed & \cmark & \cmark & \cmark & Text + 3D graph \\

LayoutGPT~\cite{feng2024layoutgpt}     
& \cmark & \cmark & 3D-FUTURE \cite{fu20213d} & \xmark & \xmark & \xmark & Text \\

Holodeck~\cite{yang2024holodeck}       
& \cmark & \cmark & Mixed & \xmark & \xmark & \cmark & Text \\

I-Design~\cite{ccelen2024design}       
& \cmark & \cmark & Objaverse \cite{deitke2023objaverse} & \cmark & \cmark & \cmark & Text \\

LayoutVLM~\cite{sun2025layoutvlm}      
& \cmark & \cmark & Objaverse \cite{deitke2023objaverse} & \xmark & \xmark & \cmark & Text + Image + 3D \\

Programmable-Room~\cite{yang2025sceneweaver}     
& \xmark & \cmark & 3D-FUTURE \cite{fu20213d} & \cmark & \xmark & \xmark & Text \\

SceneWeaver~\cite{yang2025sceneweaver}     
& \xmark & \xmark & Mixed & \xmark & \xmark & \cmark & Text \\

HSM ~\cite{pun2025hsm}     
& \xmark & \cmark & HSSD-200 \cite{khanna2024habitat} & \cmark & \cmark & \cmark & Text \\

\midrule
\textbf{RoomPilot (Ours)}              
& \textbf{\cmark} 
& \textbf{\cmark} 
& \textbf{Mixed + Collected}
& \textbf{\cmark} 
& \textbf{\cmark} 
& \textbf{\cmark} 
& Text + CAD floor plan\\
\bottomrule
\end{tabular}
}
\label{tab:systemcomparison}
\vspace{-0.8em}
\end{table*}

3D scene synthesis has widespread applications in fields such as embodied artificial intelligence, robot training and augmented reality. 
Real-world applications require scenes to be task-ready rather than merely visually plausible: robots navigate multi-room layouts and interact with objects, and AR/VR systems demand structurally consistent spaces.
Consequently, practical scene synthesis systems aim to faithfully respond to user instructions and offer explicit control over scene structure and spatial relations to ensure both usability and physical plausibility.

% Existing methods for indoor scene generation fall into three categories. 
% Procedural generation techniques \cite{deitke2022, infinigen2024indoors} create functional environments through predefined rules but lack diversity and cannot interpret natural language inputs.
% Neural generation methods \cite{tang2023diffuscene, lin2024instructscene} enable high-quality single-room synthesis with text control but are limited to isolated rooms and fail to generate cohesive multi-room layouts. 
% NeRF-based methods \cite{hollein2023text2room} can produce large-scale environments but suffer from geometric inconsistencies and accumulated structural drift.
% Beyond these individual limitations, a system-level comparison (Table~\ref{tab:systemcomparison}) reveals deeper structural gaps across existing approaches. Most methods support only partial controllability, offer limited or no interactivity, and lack compatibility with CAD tools. Their representations—typically scene graphs, object tokens, or implicit neural features—are not designed to encode architectural logic or multi-level functional relationships, making it difficult to ensure coherent multi-room layouts or practically usable spatial semantics. As a result, generated scenes often remain static, non-operable, and unsuitable for embodied AI applications that require object-level functionality.

Despite steady progress in indoor scene synthesis, existing approaches still fall short of holistically satisfying the combined requirements of multi-room support, fine-grained controllability, explicit spatial relations, and physical plausibility, as summarized in Table~\ref{tab:systemcomparison} and Figure \ref{fig:teaser}.
Model-based methods \cite{paschalidou2021atiss,tang2023diffuscene, yang2024physcene} demonstrate strong data-driven layout generation capabilities, but generalize poorly to novel scenes due to the scarcity of high-quality 3D datasets.
Rule-based and procedural systems \cite{infinigen2024indoors, raistrick2024infinigen, procthor, dai2024acdc} can produce physically valid multi-room scenes, but their rigid specifications or perception-centric designs offer limited controllability and weak support for detailed textual descriptions.
Vision-based approaches \cite{yu2025metascenes, wangarchitect} combine multiple sources of structure and semantics, enabling multi-room synthesis and relational reasoning, but visual features are difficult to precisely control scene layout.
More recently, LLM-based methods \cite{feng2024layoutgpt, ccelen2024design, sun2025layoutvlm, yang2024holodeck, yang2025sceneweaver, Kim2025Programmable} significantly enhance open-vocabulary understanding and user-level controllability, but provide limited control over spatial relations between objects.
% Hybrid agentic systems \cite{yang2025sceneweaver, Kim2025Programmable} improve physical plausibility through iterative reasoning and execution, yet still operate at the single-room level and provide limited explicit control over spatial relations and detailed scene specifications.
Collectively, these limitations indicate that no existing paradigm alone can simultaneously meet the requirements of multi-room synthesis, relation-aware controllability, and physical plausibility under rich multimodal inputs, thereby hindering practical deployment in real-world applications.
This motivates the need for a comprehensive scene synthesis framework that supports different input modalities, aligns precisely with user-specified needs, and high-quality, physically plausible 3D scenes synthesis.

A key obstacle to achieving such a framework is the lack of a unified scene representation that provides greater controllability over scene synthesis.
% Existing methods have made substantial efforts to enhance user controllability in 3D indoor scene synthesis, exploring a wide range of strategies to better 3D scene synthesis outcomes with user intent.
Many approaches rely on implicit or pipeline-specific representations: layout-driven methods encode scenes through object bounding boxes or spatial relations \cite{paschalidou2021atiss,yang2021layouttransformer, hollein2023text2room, tang2023diffuscene, ccelen2024design,jiang2024scenediffuser}, while procedural or data-driven systems adopt hand-crafted or simulator-specific abstractions to enable large-scale scene generation \cite{procthor, raistrick2024infinigen, infinigen2024indoors}. 
% Infinigen Indoors focuses on diversity and scalability via procedural rules and stochastic programs, where scene structure is implicitly embedded in generation scripts rather than explicitly exposed as a semantic representation. 
Meanwhile, Large Language Models (LLMs) methods coordinate multi-stage generation through prompt templates and tool-dependent schemas \cite{schult2024controlroom3d, yang2024holodeck, yang2025sceneweaver, Kim2025Programmable}, coupling scene structure tightly with specific synthesis modules. 
Although effective within their respective paradigms, these representations are either implicit, tightly bound to generation logic, or fragmented across stages, making it difficult to consistently reason about scene geometry, layout, structure, and semantics in a single abstraction.

To address these challenges, we propose RoomPilot, a unified framework for controllable indoor scene synthesis from detailed textual descriptions or CAD floor plans.
Inspired  by \cite{infinigen2024indoors}, RoomPilot proposes an Indoor Domain-Specific Language (IDSL), a structured intermediate representation that translates heterogeneous inputs into a coherent, multi-level semantic specification. IDSL establishes an interpretable language specification between user intent and indoor scenes, providing greater controllability over 3D scenes synthesis.
Building on IDSL, RoomPilot comprises three tightly integrated components: 
(1) Cross-modal Semantic Parsing, which converts text and CAD inputs into an initial IDSL using LLM; 
(2) Self-Regulating Scene Optimization, which iteratively refines layouts by balancing structural stability and semantic alignment, enabling coherent multi-room layouts; and 
(3) Hierarchical Indoor Scene Generation, which instantiates optimized IDSL into high-fidelity scenes with our annotated asset dataset and public datasets.
RoomPilot enables an controllable 3D scene synthesis process in which global layout, room organization, and object relations are jointly optimized under explicit structural and semantic constraints. 
Extensive experiments demonstrate that RoomPilot achieves strong controllability, physical plausibility, and visual–semantic alignment, while ablation studies confirm that both IDSL and the self-regulating optimization are essential to high-quality scene synthesis.
In summary, our contributions are as follows:
\begin{itemize}
    \item We present RoomPilot, a unified framework for controllable indoor scene synthesis from textual descriptions or CAD floor plans, enabling coherent multi-room synthesis with explicit structural and semantic control.
    \item RoomPilot introduces the \textit{Indoor Domain-specific Language (IDSL)}, a unified intermediate representation that translates multi-modal inputs into structured semantic and geometric design specifications, providing an interpretable and controllable bridge from user intentions to 3D scene synthesis.
    \item We propose a self-regulating, energy-based hierarchical generation algorithm that progressively organizes scenes from three sematic levels, enabling coherent multi-room layouts with consistent structural organization and semantic alignment. We construct a curated annotated asset dataset, which supports synthesizing high-fidelity 3D scenes.
\end{itemize}

\section{Related Work}
\label{sec:Related Work}

\subsection{Procedural Based Scene Generation}
Procedural Content Generation (PCG) has seen extensive research for generating both outdoor and indoor 3D scenes \cite{procthor, infinigen2024indoors, raistrick2024infinigen, zhou2025roomcraft, gasch2022procedural, zhou2024scenex, sun20253d}.
Infinigen Indoors \cite{infinigen2024indoors} is capable of generating infinite indoor scenes.
RoomCraft \cite{zhou2025roomcraft} generates complete indoor scenes from multiple modalities. 
ProcTHOR \cite{procthor} supports infinite interactive indoor scene generation, yet lacks user-driven customization.
However, these methods typically rely on hand-crafted rules or programmatic specifications, where scene structure and object relationships are implicitly encoded in generation scripts.
As a result, while procedural approaches excel at scalability and physical plausibility, they offer limited controllability at the semantic and relational level, making it difficult to precisely align generated scenes with user intentions.

\subsection{Neural Single-Room Generation}
Recent works have explored using neural models for single-room generation \cite{paschalidou2021atiss, tang2023diffuscene, lin2024instructscene, feng2024layoutgpt, schult2024controlroom3d}.
DiffuScene \cite{tang2023diffuscene} applies diffusion on unordered object sets for diverse and realistic 3D scene synthesis while InstructScene \cite{lin2024instructscene} uses a semantic graph prior and layout decoder to generate controllable 3D scenes from language instructions. LayoutGPT \cite{feng2024layoutgpt} adopts Large Language Models (LLMs) as visual planners to generate structured 2D/3D layouts.
These approaches are effective for single rooms but are not designed to model global structure, inter-room relations, or architectural constraints required for coherent multi-room scene generation.
% These models generate visually realistic scenes with precise text-based control. However, they are restricted to single-room layouts, lack integration with CAD data and asset retrieval mechanisms, and exhibit limited scalability to complex, multi-room environments.

\subsection{Large-Scale Scene Generation}
A recent line of work has focused on generating large-scale scenes \cite{hollein2023text2room, hu2024scenecraftllmagentsynthesizing, deng2023mv, zhang2024furniscene, yang2024scenecraft, fang2025ctrl}.
Ctrl-Room \cite{fang2025ctrl} and SceneCraft \cite{yang2024scenecraft} generate 3D room meshes using layout guidance, while Ctrl-Room \cite{fang2025ctrl} using layout diffusion and panoramic NeRFs, and SceneCraft \cite{yang2024scenecraft} combining bounding-box layouts with 2D diffusion and distillation.
Text2Room \cite{hollein2023text2room} generates textured 3D room meshes by iteratively fusing multi-view images synthesized from text prompts using inpainting and monocular depth estimation.
However, these methods rely on implicit or image-centric representations to scale scene generation.
As a result, maintaining explicit room-level semantics, and controllable object relations across complex multi-room environments remains challenging.

\subsection{Intermediate Representations and Constraint Optimization}
Recent advances in scene generation have explored intermediate representations such as scene graphs \cite{ccelen2024design, lin2024instructscene, tang2023diffuscene, bai2023componerf} and layout optimization \cite{paschalidou2021atiss, fang2025ctrl, yang2024scenecraft, feng2024layoutgpt, ocal2024sceneteller, yang2024holodeck}. 
I-Design \cite{ccelen2024design} employs LLM agents to convert free-form text into scene graphs, optimize object layouts, and retrieve assets for personalized 3D interiors. ATISS \cite{paschalidou2021atiss} models scene synthesis as autoregressive set generation with transformers, while SceneTeller \cite{ocal2024sceneteller} uses in-context learning and 3D Gaussian Splatting to produce controllable and consistent scenes.
These methods encode object relations and spatial constraints using representations at specific abstraction levels.
But the lack of a unified representation across architectural, layout, and object levels limits holistic expressing in complex indoor scenes.

% In contrast, our IDSL offers a unified representation that is more closely aligned with the design process, providing greater flexibility for generating complex, multimodal indoor environments.
% IDSL unifies multimodal input parsing, constraint modeling, and hierarchical generation. It achieves, for the first time, end-to-end generation from CAD drawings to fully interactive suites, offering a comprehensive solution for controlled scene creation.

\section{Method}
\label{sec:Method}

\begin{figure*}[hbtp]
    \centering
    % \vspace{-0.8em}
    \includegraphics[width=1\linewidth]{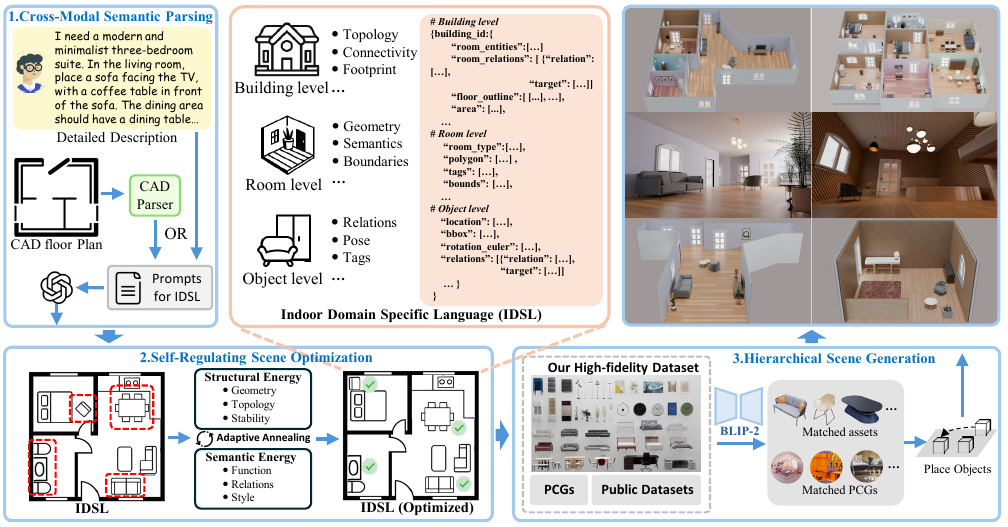}
    \caption{\textbf{RoomPilot} takes either text descriptions or CAD floor plans as input, parses them into a unified IDSL representation, optimizes multi-room layouts by a self-regulating energy-based process, and retrieves appropriate assets to generate high-fidelity 3D indoor scenes. 
    }
    \label{fig:overview}
    \vspace{-1.3em}
\end{figure*}

% \begin{figure}[hbtp]
%    \centering
%   \includegraphics[width=0.8\linewidth]{Figures/Fig_IDSL.pdf}
%   \captionof{figure}{The structure of IDSL. }
%   \label{fig:idsl}
% 	\vspace{-20pt}
% \end{figure}

% 给定用户的自然语言描述T和CAD户型图F，我们的目标是生成一个完整的3D室内场景S，
% 包含建筑结构（墙体、门窗）和家具布置，同时满足几何一致性、语义合理性和交互性要求。
% 我们的方法包含三个阶段（如图2所示）：
% (1) Cross-Modal Semantic Parsing: 将T和F分别通过LLM和几何解析器转换为IDSL表示；
% (2) Self-Regulating Scene Optimization:采用基于退火的自调节能量演化机制，在几何可行性与语义一致性之间动态调控平衡，对 IDSL 中对象的空间布局与交互关系进行全局优化。 
% (3) Hierarchical Scene Generation: 基于优化后的IDSL生成建筑结构、家具及其空间关系；
% 关键创新在于IDSL作为统一接口，解耦了设计意图的表达与具体实现，使得每个阶段都可以独立优化同时保持全局一致性。

Given a detailed textual description $T$ or a CAD floor plan $F$ as the sole input, our goal is to generate a complete 3D indoor scene $S$.
% that integrates architectural structures (walls, doors, windows) and furniture arrangements while ensuring geometric consistency, semantic coherence. 
As illustrated in Figure \ref{fig:overview}, our framework consists of three modules:
(1) \textbf{Cross-Modal Semantic Parsing}: Textual descriptions or CAD floor plans are transformed into an initial IDSL configuration $\mathcal{S}_0$ using an LLM.
(2) \textbf{Self-Regulating Scene Optimization}: Starting from the initial IDSL configuration $\mathcal{S}_0$, the scene is iteratively refined by jointly considering structural stability and semantic alignment to get the optimized IDSL $\mathcal{S}^\ast$.
(3) \textbf{Hierarchical Scene Generation}: The optimized IDSL $\mathcal{S}^\ast$ is instantiated into a complete indoor scene with architectural structures, furniture assets, and their spatial relationships.
% The key innovation lies in using IDSL as a unified semantic interface, which decouples the expression of design intent from its physical realization, allowing each stage to be optimized independently while maintaining global scene consistency.

% Given a natural language description \( T \) and a CAD floor plan \( F \), our goal is to generate a complete 3D indoor scene \( S \) that incorporates architectural structures (walls, doors, and windows) and furniture arrangements, while ensuring geometric consistency, semantic plausibility, and interactivity. As illustrated in Fig.~\ref{fig:overview}, our framework consists of four stages: (1) \textbf{Multi-modal Parsing}, which converts \( T \) and \( F \) into IDSL representation using LLMs and a geometric parser; (2) \textbf{Constraint Optimization}, which refines the spatial relationships among objects within IDSL through simulated annealing; (3) \textbf{Structure Generation}, which constructs the architectural layout (walls, doors, and windows) based on the optimized IDSL; and (4) \textbf{Asset Population}, which retrieves and places interactive furniture assets from a large-scale database. The key innovation lies in employing IDSL as a unified interface that decouples design intent from implementation, enabling each stage to be independently optimized while maintaining global coherence across the entire generation pipeline.

% \begin{figure}[hbtp]
%     \centering
%     \includegraphics[width=1\linewidth]{Figures/Fig_IDSL.pdf}
%     \caption{.}
%     \label{fig:idsl}
% \end{figure}

\subsection{Cross-Modal Semantic Parsing}
\label{subsec:parsing}

\subsubsection{Text Description Parsing.}
\label{subsubsec:Text Description Parsing}
Given a textual scene description $T$, we adopt a hierarchical parsing strategy aligned with the three-level structure of IDSL. We first prompt GPT-4 with $T$ and the IDSL building-level schema to extract global scene structure: room entities, types, inter-room topology, room contours with consistent shared edges, and structural openings (doors, windows, open passages) positioned on shared or exterior walls. The prompt includes IDSL schema definitions as output format constraints and few-shot examples. During this step, the input description is also segmented into per-room textual fragments $\{T_1, \dots, T_K\}$. Soft preferences expressed in the description (e.g., relative size requirements) are resolved by the LLM during contour generation and reflected in the resulting geometry.

For each room $k$, we invoke GPT-4 with building-level context (including fixed room contours and opening positions) and room-specific description $T_k$ to extract object entities, attributes, and intra-room relations, producing a per-room IDSL configuration. Since room geometry and structural openings are fully determined in the building-level pass, each room-level invocation focuses solely on object placement within a fixed boundary, avoiding geometric inconsistencies across rooms. Processing rooms independently with shared building-level context also avoids LLM memory bottlenecks in multi-room scenarios.

The assembled configurations form the initial IDSL $\mathcal{S}_0$. A rule-based validator checks schema compliance and topological consistency; remaining semantic inconsistencies are resolved by downstream optimization (Section~\ref{subsec:optimization}). Note that the positions and bounding boxes generated at this stage serve as initial estimates; the self-regulating optimization (Section~\ref{subsec:optimization}) iteratively refines them to achieve physically plausible and semantically coherent layouts. The complete prompt templates are provided in Appendix~\ref{app:prompt-design}, and a full text-to-IDSL example is provided in Appendix~\ref{app:parsing-example}.

% \subsubsection{CAD Floor Plan Parsing}

% CAD floor plans (DXF/DWG) are processed by a geometric parser that identifies room boundaries, dimensions and openings symbols, which are directly mapped into the unified IDSL configuration $\mathcal{S}_0$. In the CAD-based setting, geometric properties (room dimensions, wall positions, opening locations) are fully determined by the floor plan, serving as hard architectural constraints for scene generation. 
% Semantic properties such as room type and furniture requirements are inferred from the geometric context (e.g., room size and connectivity) using rule-based heuristics. This design ensures structural correctness while preserving semantic coherence without requiring additional textual input.
% %
% See appendix \ref{sec:parsing_supp} for more details about cross-modal semantic parsing.

\subsubsection{CAD Floor Plan Parsing.}
\label{subsubsec:CAD Floor Plan Parsing}
For CAD floor plan, we ues a geometric parsing pipeline that extracts room contours, doors, and windows from the vectorized representation (e.g., SVG files following the format of CubiCasa5K~\cite{kalervo2019cubicasa5k}). The input floor plan contains semantically grouped elements labeled as \texttt{Space}, \texttt{Door}, and \texttt{Window}.

For each \texttt{Space} group, the parser extracts polygon vertices, applies the SVG transform chain to obtain world coordinates, and selects the largest polygon by area as the room contour. A global vertex snapping step clusters nearby vertices within a distance threshold and replaces them with cluster centroids, ensuring adjacent rooms share exact wall coordinates. Shared edges between rooms are computed via polygon boundary intersection. For doors and windows, the parser computes the centroid of each group in world coordinates and matches it to the two nearest room boundaries by point-to-segment distance.
% , establishing \texttt{CutFrom} relations in IDSL. 
Room types are extracted from the class attributes of each \texttt{Space} group, and inter-room adjacency is derived from the shared-edge computation.

Since CAD floor plans provide precise room geometry but typically lack detailed furniture layouts, object-level content is generated by invoking the room-level LLM prompt (Section~\ref{subsubsec:Text Description Parsing}), with the room description replaced by the room type as the generation cue. The LLM infers typical furniture configurations based on room function (e.g., a bedroom typically contains a bed, nightstands, and a wardrobe), unifying both input modalities through IDSL. The complete geometric parsing procedure is detailed in Appendix~\ref{app:cad-parsing}.

\subsection{IDSL Representation}
\label{subsec:idsl} 

% 我们提出了IDSL，这是第一个将室内场景生成看作一个编译问题的领域特定语言——把高级设计规范转化成可执行的场景配置。与现有的表示方法不同，roompilot强行在抽象（scene graph）和具体（体素，bounding box）之前去制造了二分法，IDSL在一个中间语义层级上操作，这个层级既具有计算上的可处理性，又能表达设计意图。
% 把创新凝练成，IDSL的创新如下：（统一的多模态接口）作为通用异构输入的通用中间表示；基于约束的设计：结构化的语义、几何和约束规范，消除了后期约束注入的需求；双向的可编辑性，人类可读的语法，使得它既可以进行自动生成，也能进行手动编辑。
% To overcome these challenges, IDSL is the first domain-specific language that treats indoor scene synthesis as a compilation problem, translating high-level design specifications into executable scene configurations. 
% To overcome these challenges, we propose IDSL, a domain-specific language that treats indoor scene synthesis as a compilation problem, translating high-level design specifications into executable scene configurations.
Unlike existing representations that impose a rigid dichotomy between highly abstract forms (e.g., scene graphs) and overly concrete geometric encodings (e.g., voxels or bounding boxes), 
IDSL provides a unified interface for heterogeneous inputs as a structured intermediate representation between user intent and indoor scenes, which establishes interpretable semantic and geometric specifications to enable greater controllability over indoor scene synthesis. 
% IDSL establishes interpretable semantic and geometric specifications, thus enabling greater controllability over indoor scene synthesis.
We provide a detailed discussion on IDSL and its comparison to Infinigen Indoors in Appendix \ref{sec:idsl_supp}.
% （1）在足够抽象的层次上统一多模态输入，如自然语言和 CAD 平面图，以便实现统一的中间表征；（2）在具体化的层面上提供明确的指导，以参数化形式达到可控，以便有效生成和优化空间布局和空间位置关系。
% IDSL中，一定要有图片展示例子，且在正文里有提到让看补充材料（如果正文放不下）

% we introduce the Indoor Domain-Specific Language (IDSL), which aims to establish a balanced representation between abstraction and concreteness. Specifically, IDSL is designed to: (1) be sufficiently abstract to unify multi-modal inputs such as natural lauguage and CAD floor plans; (2) be sufficiently concrete to guide the generation and optimization of indoor scenes; and (3) remain human-readable and editable to seamlessly integrate into professional design workflows. The outputs of cross-modal parsing (Section~\ref{subsec:parsing}) are normalized into this language, which then serves as the input state for the optimization stage.

% \textbf{Multi-Level Semantic Hierarchy.} IDSL is a hierarchical and formal modeling framework for precise and controllable representation of indoor scenes. It is grounded in constraint logic programming and adopts a declarative syntax to enable structured and interpretable modeling of complex spatial layouts. IDSL defines a three-level semantic hierarchy—Building Level, Room Level, and Object Level—each responsible for different aspects of spatial organization and semantic description.

\paragraph{Multi-Level Semantic Hierarchy.}
IDSL is a hierarchical modeling framework for indoor scenes that enables precise and controllable scene representation. It uses a declarative, rule-based formulation to explicitly describe spatial structures and constraints in an interpretable manner.
IDSL defines a three semantic hierarchy—Building Level, Room Level, and Object Level, each deliberately designed according to architectural principles. 
% The Building Level ensures structural completeness and global spatial topology; the Room Level captures functional zoning and intra-room organization; and the Object Level governs fine-grained furniture configuration. 
Each level corresponds to distinct design objectives and optimization rules that are difficult to merge or further subdivide, making the three-level structure both necessary and well-aligned with practical indoor scene design workflows.

% 解释为什么要用三个层次？（三个层次是基于建筑设计的原理精心设计的:在建筑层级，用于区保证结构的完整性，房间层级保证功能分区，物体层级去用于家具配置。每个层级对应不同的优化目标和优化规则，这些目标无法有效地合并和进一步地细分）
% 

At the \textit{Building Level}, IDSL includes the global spatial topology and geometric configuration. 
It models inter-room connectivity through an adjacency list, where nodes denote individual spatial units and edges signify physical adjacencies. 
The geometric boundaries are defined by two-dimensional vertex sequences, enabling the precise representation of concave and complex room footprints. 
% Furthermore, structural constraints—including validated wall segments and the semantic integration of openings like doors and windows—ensure architectural feasibility.

At the \textit{Room Level}, IDSL focuses on spatial semantics and functional intent.
Each room is described by its semantic and basic structural properties. This level defines how design requirements are represented in a structured form, so that room functions, layout expectations, and object organization are clearly specified rather than implicitly assumed. As a result, the system not only identifies the type of room, but also constrains how it should be arranged.

At the \textit{Object Level}, IDSL provides the finest granularity of modeling.
This level defines objects through a structured set of semantic categories, geometric attributes, and explicit relations, allowing consistent specification of object properties while preserving flexibility in spatial configuration.

\subsection{Self-Regulating Scene Optimization}
\label{subsec:optimization}

Most scene-optimization frameworks formulate layout synthesis as energy minimization with manually tuned penalty weights, entangling geometric feasibility and semantic intent in an opaque objective \cite{yu2011make, infinigen2024indoors}. We instead cast scene generation as a \emph{self-regulating} energy evolution process grounded in IDSL, where optimization priorities and exploration behavior are adaptively modulated by rule-satisfaction feedback rather than fixed schedules. Given an initial IDSL configuration $\mathcal{S}_0$ (Section~\ref{subsec:parsing}) and its hierarchical rule specification (Section~\ref{subsec:idsl}), the optimizer progressively refines the scene into a final configuration $\mathcal{S}^*$.

\paragraph{Dual-Channel Energy Design.}
We define a time-dependent total energy that explicitly decomposes the objective into a structural channel and a semantic channel:
\begin{equation}
  E(\mathcal{S}, t) \;=\; \alpha_{\mathrm{struct}}(t)\, E_{\mathrm{struct}}(\mathcal{S}) \;+\; \alpha_{\mathrm{sem}}(t)\, E_{\mathrm{sem}}(\mathcal{S}),
  \label{eq:total-energy}
\end{equation}
where $E_{\mathrm{struct}}(\mathcal{S}) = \sum_i w_i^{\mathrm{struct}}\, f_i^{\mathrm{struct}}(\mathcal{S})$ aggregates structural factors (collision penalties, boundary violations, contact and support stability), and $E_{\mathrm{sem}}(\mathcal{S}) = \sum_j w_j^{\mathrm{sem}}\, f_j^{\mathrm{sem}}(\mathcal{S})$ captures semantic factors (functional proximity, directional alignment, relational satisfaction). Here $w_i^{\mathrm{struct}}$ and $w_j^{\mathrm{sem}}$ are pre-defined importance weights for each factor. Both terms yield scalar penalties proportional to degree of constraint violation. The channel weights $\alpha_{\mathrm{struct}}(t)$ and $\alpha_{\mathrm{sem}}(t)$ are adaptively adjusted based on rule-alignment score $\rho(t)$ (defined below): as structural constraints become satisfied, $\alpha_{\mathrm{struct}}(t)$ decreases and $\alpha_{\mathrm{sem}}(t)$ increases, progressively shifting optimization focus from geometric feasibility to semantic organization.

\paragraph{Adaptive Annealing with Self-Feedback.}
At each iteration, a candidate configuration $\mathcal{S}'$ is generated by applying a transformation operator $\Omega$ (translation, rotation, swapping, or group-level adjustment). The key self-regulating mechanism is driven by a rule-alignment score $\rho(t) \in [0,1]$ that measures normalized satisfaction of active IDSL factors. This score simultaneously controls two aspects of the search:

\noindent\textit{Operator selection.}\; Operators are sampled from a time-varying distribution:
\begin{equation}
  p_i(t) \;=\; \frac{\beta_i(t)}{\sum_j \beta_j(t)}, \quad
  \beta_i(t) \;=\; \beta_i^{(0)}\,(1 - \rho(t))^{\gamma_i},
  \label{eq:operator}
\end{equation}
where $\beta_i^{(0)}$ is the initial weight and $\gamma_i$ controls the decay rate of operator $i$. Different operators are assigned distinct $\gamma_i$ values so that disruptive operators (\emph{e.g.}, swapping) decay faster than local refinements (\emph{e.g.}, small translations), ensuring that large-scale rearrangements dominate early exploration while fine adjustments prevail near convergence.

\noindent\textit{Stochastic temperature.}\; The annealing temperature is coupled to the same feedback signal:
\begin{equation}
  T(t) \;=\; T_0\,(1 - \rho(t))^{\delta},
  \label{eq:temperature}
\end{equation}
where $T_0$ is the initial temperature and $\delta$ controls the decay shape. Exploration intensity decreases as constraints become satisfied, without requiring a hand-crafted cooling schedule.

Proposal acceptance follows a dual-channel Metropolis criterion: any modification that reduces $E_{\mathrm{struct}}$ is always accepted, ensuring monotonic convergence toward geometric feasibility. Changes that preserve structural soundness but affect semantic organization are accepted with probability $\exp(-\Delta E_{\mathrm{sem}} / T(t))$, allowing semantic refinement without reintroducing structural violations.

\paragraph{Progressive Rule Activation.}
To prevent premature coupling of heterogeneous objectives, IDSL rules are partitioned into ordered groups $\{\mathcal{R}^{(1)}, \dots, \mathcal{R}^{(L)}\}$ corresponding to structural, functional, and stylistic reasoning levels. Rules are activated incrementally: once $\rho(t)$ for the current active rules saturates and local energy variation diminishes, the next group of rules is activated. This produces a coarse-to-fine optimization trajectory aligned with the IDSL hierarchy.

\paragraph{Localized Evaluation and Convergence.}
To maintain scalability, energy changes are evaluated locally over modified objects $\mathcal{N}$ and their spatial neighborhood $\mathrm{adj}(\mathcal{N})$:
\begin{equation}
  \Delta E \;\approx\; \sum_{r \,\in\, \mathrm{adj}(\mathcal{N})} \bigl( E'_r - E_r \bigr),
  \label{eq:local-eval}
\end{equation}
where $r$ indexes affected IDSL factors. The optimization terminates when both structural drift and semantic variation plateau and further rule activation yields negligible energy reduction, producing a structurally feasible and semantically coherent configuration $\mathcal{S}^*$.

\begin{table*}[t]
\centering
\caption{\textbf{Quantitative comparison} between RoomPilot and existing scene synthesis methods based on detailed text instructions.}
\vspace{-0.3em}
\resizebox{\textwidth}{!}{
\begin{tabular}{lccccccc|ccccccc|ccccccc}
\toprule
\multirow{3}{*}{\textbf{Method}} &
\multicolumn{7}{c|}{\textbf{Bedroom}} &
\multicolumn{7}{c|}{\textbf{Living Room}} &
\multicolumn{7}{c}{\textbf{Dining Room}} \\ 
\cmidrule(lr){2-8} \cmidrule(lr){9-15} \cmidrule(lr){16-22}
 & \multicolumn{3}{c}{\textbf{Physics}} 
 & \multicolumn{4}{c|}{\textbf{Visual \& Semantics}} 
 & \multicolumn{3}{c}{\textbf{Physics}} 
 & \multicolumn{4}{c|}{\textbf{Visual \& Semantics}} 
 & \multicolumn{3}{c}{\textbf{Physics}} 
 & \multicolumn{4}{c}{\textbf{Visual \& Semantics}} \\
 & \#Obj $\uparrow$ & \#OB $\downarrow$ & \#CN $\downarrow$
 & Real. $\uparrow$ & Func. $\uparrow$ & Lay. $\uparrow$ & Comp. $\uparrow$
 & \#Obj $\uparrow$ & \#OB $\downarrow$ & \#CN $\downarrow$
 & Real. $\uparrow$ & Func. $\uparrow$ & Lay. $\uparrow$ & Comp. $\uparrow$
 & \#Obj $\uparrow$ & \#OB $\downarrow$ & \#CN $\downarrow$
 & Real. $\uparrow$ & Func. $\uparrow$ & Lay. $\uparrow$ & Comp. $\uparrow$ \\
\midrule
LayoutGPT \cite{feng2024layoutgpt}
 & 5.4 & 0.4 & 3.8 & 6.1 & 4.6 & 5.3 & 4.8 
 & 6.5 & 1.6 & 4.5 & 4.2 & 3.0 & 2.4 & 3.4 
 & 8.6 & 0.1 & 3.1 & 4.2 & 5.3 & 3.3 & 5.3 \\
Holodeck \cite{yang2024holodeck}
 & \textbf{14.9} & 1.5 & 0.1 & 8.6 & 5.8 & 6.6 & 5.1  
 & 13.2 & 1.1 & 0.3 & 8.3 & 8.8 & 7.1 & 3.8  
 & 12.3 & 2.0 & 0.1 & 8.5 & 8.6 & 6.7 & 4.2 \\
I-Design \cite{ccelen2024design}
 & 9.0 & 2.8 & 2.1 & 3.3 & 2.5 & 4.0 & 3.5 
 & 19.8 & 4.8 & 10.2 & 8.6 & 7.6 & 7.7 & 7.7 
 & 19.2 & 3.5 & 10.5 & 4.9 & 4.3 & 5.7 & 3.8 \\
SceneTeller \cite{ocal2024sceneteller}
 & 6.2 & 0.2 & 3.0 & 5.8 & 6.5 & 5.9 & 3.5   
 & 6.8 & 0.1 & 1.7 & 6.0 & 6.6 & 4.1 & 3.8  
 & 7.9 & 0.0 & 2.1 & 5.5 & 6.4 & 4.6 & 4.0  \\ 
LayoutVLM \cite{sun2024layoutvlm}
 & 6.6 & 0.5 & 0.0 & 4.5 & 3.0 & 3.3 & 3.0
 & 5.3 & 0.3 & 0.5 & 5.2 & 2.8 & 3.5 & 4.6 
 & 5.0 & 0.2 & 0.3 & 4.8 & 6.3 & 4.0 & 3.8 \\
SceneWeaver \cite{yang2025sceneweaver}
 & 10.1 & 0.1 & 0.1 & 7.1 & 6.9 & 6.8 & 6.2
 & 15.4 & 0.0 & 0.2 & 7.6 & 7.8 & 5.9 & 6.1 
 & 13.8 & 0.0 & 0.0 & 5.9 & \textbf{7.1} & 5.8 & 4.8 \\
HSM~\cite{pun2025hsm}
 & 10.6 & 0.1 & 0.1 & 8.9 & 9.0 & 7.6 & 7.8
 & 20.4 & 0.0 & 0.1 & 7.7 & 8.0 & 7.0 & 7.1
 & 15.7 & 0.0 & 0.1 & 5.4 & 5.9 & 4.9 & 4.6 \\
\midrule
\textbf{RoomPilot}
 & 13.9 & \textbf{0.0} & \textbf{0.0} & \textbf{9.1} & \textbf{9.5} & \textbf{8.2} & \textbf{9.2}
 & \textbf{25.8} & \textbf{0.0} & \textbf{0.0} & \textbf{8.8} & \textbf{9.2} & \textbf{7.7} & \textbf{8.4}
 & \textbf{20.9} & \textbf{0.0} & \textbf{0.0} & \textbf{6.3} & 6.4 & \textbf{5.7} & \textbf{5.4} \\
\bottomrule
\multirow{3}{*}{\textbf{Method}} &
\multicolumn{7}{c|}{\textbf{Kitchen}} &
\multicolumn{7}{c|}{\textbf{Bathroom}} &
\multicolumn{7}{c}{\textbf{Average}} \\
\cmidrule(lr){2-8} \cmidrule(lr){9-15} \cmidrule(lr){16-22}
 & \multicolumn{3}{c}{\textbf{Physics}} 
 & \multicolumn{4}{c|}{\textbf{Visual \& Semantics}} 
 & \multicolumn{3}{c}{\textbf{Physics}} 
 & \multicolumn{4}{c|}{\textbf{Visual \& Semantics}} 
 & \multicolumn{3}{c}{\textbf{Physics}} 
 & \multicolumn{4}{c}{\textbf{Visual \& Semantics}} \\
 & \#Obj $\uparrow$ & \#OB $\downarrow$ & \#CN $\downarrow$
 & Real. $\uparrow$ & Func. $\uparrow$ & Lay. $\uparrow$ & Comp. $\uparrow$
 & \#Obj $\uparrow$ & \#OB $\downarrow$ & \#CN $\downarrow$
 & Real. $\uparrow$ & Func. $\uparrow$ & Lay. $\uparrow$ & Comp. $\uparrow$
 & \#Obj $\uparrow$ & \#OB $\downarrow$ & \#CN $\downarrow$
 & Real. $\uparrow$ & Func. $\uparrow$ & Lay. $\uparrow$ & Comp. $\uparrow$ \\
\midrule
LayoutGPT \cite{feng2024layoutgpt}
 & 4.9 & 0.4 & 0.1 & 4.3 & 5.3 & 5.6 & 2.2 
 & 4.3 & 1.1 & 2.2 & 4.3 & 5.5 & 3.3 & 4.6 
 & 5.9 & 0.7 & 2.7 & 5.0 & 4.7 & 4.0 & 3.9 \\
Holodeck \cite{yang2024holodeck}
 & 8.6 & 1.8 & 0.1 & 4.2 & 4.3 & 2.3 & 3.6 
 & 9.8 & 1.9 & 0.1 & 2.9 & 4.9 & 4.2 & 2.7 
 & 11.8 & 1.7 & 0.1 & 6.5 & 6.5 & 5.0 & 3.9 \\
I-Design \cite{ccelen2024design}
 & \textbf{21.6} & 4.1 & 10.4 & 3.3 & 4.6 & 4.8 & 4.5 
 & \textbf{14.2} & 5.2 & 3.8 & 4.6 & 3.3 & 3.3 & 3.4 
 & \textbf{16.8} & 4.1 & 7.4 & 5.0 & 4.5 & 6.0 & 4.6 \\
SceneTeller \cite{ocal2024sceneteller}
 & 5.7 & 0.5 & 0.7 & 5.4 & 6.0 & 3.5 & 3.2  
 & 6.5 & 0.7 & 0.5 & 5.2 & 5.9 & 3.3 & 3.4 
 & 6.6 & 0.3 & 1.6 & 5.6 & 6.3 & 4.3 & 3.6 \\
LayoutVLM \cite{sun2024layoutvlm}
 & 3.6 & 0.1 & 0.1 & 5.2 & 4.4 & 3.9 & 2.3 
 & 6.3 & 0.7 & 0.1 & 4.2 & 3.3 & 2.9 & 3.3
 & 5.4 & 0.4 & 0.2 & 4.8 & 4.0 & 3.5 & 3.4 \\
SceneWeaver \cite{yang2025sceneweaver}
 & 9.1 & 0.2 & 0.2 & 6.5 & 5.9 & 5.2 & 4.6
 & 8.3 & 0.3 & 0.2 & 5.1 & 5.8 & 3.9 & 4.5 
 & 11.3 & 0.1 & 0.1 & 6.4 & 6.7 & 5.5 & 5.2 \\
HSM~\cite{pun2025hsm}
 & 11.8 & 0.1 & 0.1 & 7.4 & 5.5 & 6.5 & 4.3
 & 11.1 & 0.1 & 0.1 & 5.0 & 5.3 & 4.5 & 5.1
 & 13.9 & 0.1 & 0.1 & 6.9 & 6.7 & 6.1 & 5.8 \\
\midrule
\textbf{RoomPilot}
 & 10.2 & \textbf{0.0} & \textbf{0.0} & \textbf{7.8} & \textbf{6.6} & \textbf{6.8} & \textbf{5.2} 
 & 10.6 & \textbf{0.0} & \textbf{0.0} & \textbf{6.1} & \textbf{6.5} & \textbf{5.6} & \textbf{5.3} 
 & 16.3 & \textbf{0.0} & \textbf{0.0} & \textbf{7.6} & \textbf{7.6} & \textbf{6.8} & \textbf{6.7} \\
\bottomrule
\end{tabular}
}
\label{tab:quantitative comparison}
\vspace{-8pt}
\end{table*}

\subsection{Hierarchical Indoor Scene Generation}
\label{sec:hierarchical indoor generation}

We adopt a hierarchical generation pipeline---\textit{building}, \textit{room}, and \textit{object}---to ensure controllability, modularity, and architectural consistency (see Appendix~\ref{sec:generation_supp} for implementation details).

\textbf{Building-Level Structural Generation.}
Given room polygons $\mathcal{P}=\{P_i\}$ and adjacency graph $\mathcal{G}$ from IDSL, 
we synthesize the global architectural shell using a procedural generator:
\begin{equation}
\mathcal{W} = \text{PCG}_{\text{wall}}(\mathcal{P}, \mathcal{G}, \theta_{\text{struct}}),
\end{equation}
where $\theta_{\text{struct}}$ controls wall height, thickness, and corner styles. 
The generator follows an explicit geometric pipeline: 
(1) each polygon $P_i$ is converted into a consistently oriented closed baseline curve; 
(2) wall strips are constructed by extruding polygon edges with thickness $t$ and height $h$, 
with adaptive subdivision applied near concave corners and long edges to avoid geometric degeneracies; 
(3) adjacent wall segments are connected using corner rules (miter/butt joins) and merged via Boolean union, 
followed by mesh cleanup to produce a watertight manifold wall mesh. 
This pipeline guarantees structural consistency across rooms and avoids gaps or overlaps.

\textbf{Room-Level Opening Construction.}
For each room, functional openings (doors and windows) are instantiated by applying Boolean subtraction on wall meshes:
\begin{equation}
\mathcal{W}_i' = \mathcal{W}_i \setminus \bigcup_{j} \mathcal{O}_j,
\end{equation}
where $\mathcal{O}_j$ denotes opening volumes. 
Parametric generators then produce opening geometries (frames, panels) and materials from predefined presets, 
ensuring both geometric validity and consistent appearance.

\textbf{Object-Level Scene Population.}
We populate rooms using a hybrid asset ecosystem combining:  
(1) a curated static corpus of \textbf{9271 annotated indoor assets} across \textbf{151 categories};  
(2) \textbf{27 category-specific procedural generators};  
(3) public 3D datasets \cite{fu20213d, chang2015shapenet, deitke2023objaverse}.  

For a target object $\mathcal{S}^\ast$ defined in IDSL, candidate assets are evaluated using a multi-modal compatibility score:
\begin{equation}
\begin{aligned}
\text{score}(a, \mathcal{S}^\ast) = \ &
\lambda_{\text{sem}} \, \text{sim}_{\text{BLIP}}(f_a, f_{\mathcal{S}^\ast}) \\
& + \lambda_{\text{geo}} \, \text{IoU}(\text{bbox}_a, \text{bbox}_{\mathcal{S}^\ast}) \\
& + \lambda_{\text{style}} \, \psi(s_a, s_{\text{room}}),
\end{aligned}
\end{equation}
where semantic, geometric, and stylistic compatibility are jointly considered. 

If the retrieval confidence is below a threshold $\tau$, a procedural generator is invoked:
\begin{equation}
a_{\text{gen}} = \text{PCG}_{\text{category}(\mathcal{S}^\ast)}(\text{params}(\mathcal{S}^\ast), \xi_{\text{style}}),
\end{equation}
where each $\text{PCG}_{\text{category}}$ constructs objects from primitive components 
(e.g., cuboids, panels, and repeated structures) using category-specific assembly rules 
(e.g., \textit{table = tabletop + legs}, \textit{cabinet = frame + panels}). 
Parameters such as size, proportions, and style are derived from IDSL specifications. 
This ensures that generated assets are structurally valid, watertight, and semantically consistent.

\textbf{Relationship-Aware Refinement.}
After placement, object poses are refined using relational constraints 
$\mathcal{R}_i=\{(o_j, r_{ij}, c_{ij})\}$ defined in IDSL. 
We optimize object positions $\mathbf{p}_i$ and orientations $\mathbf{q}_i$ via:
\begin{equation}
\begin{aligned}
\min_{\{\mathbf{p}_i, \mathbf{q}_i\}} \ \ 
& \sum_{i,j} w_{ij} \, d\!\left(T_i, T_j, c_{ij}\right) \\
& + \lambda_{\text{stable}} E_{\text{physics}},
\end{aligned}
\end{equation}
where $d(\cdot)$ measures constraint violation and $E_{\text{physics}}$ enforces stability and collision avoidance. 
This step aligns relational surfaces (e.g., objects resting on supports), 
eliminates floating or intersecting artifacts, and ensures physically plausible arrangements.

\section{Experiments}
\label{sec:Experiments}

\begin{figure*}[t]
    \centering
    \includegraphics[width=1\linewidth]{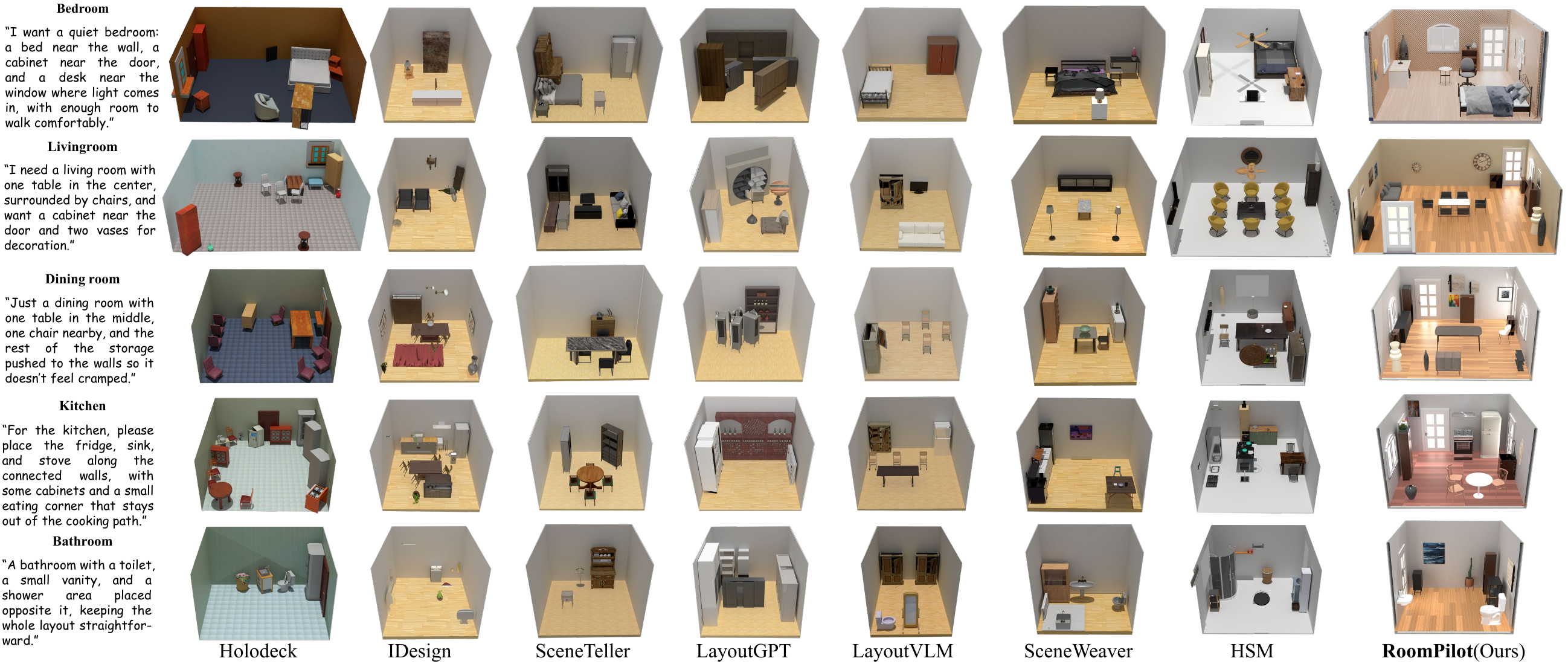}
    % \caption{\textbf{Single-room generation comparison.} We show results for five room types under identical text prompts. To ensure a fair comparison, all methods are run with their default configurations without any task-specific tuning. Methods that natively support wall materials, floor textures, doors, and windows are rendered with these features enabled; for methods that do not natively support such elements, we apply a unified default floor texture and wall material across all scenes. All scenes are rendered under consistent camera viewpoints and lighting parameters. RoomPilot produces layouts with higher semantic alignment and structural coherence compared to existing methods.}
    \caption{\textbf{Single-room generation comparison.} 
    % Results for five room types under identical text prompts. 
    All methods use default configurations without task-specific tuning.
    Methods natively supporting wall materials, floor textures, doors, and windows are rendered with these features enabled; otherwise, a unified default floor and wall material is applied. 
    All scenes share consistent camera viewpoints and lighting. RoomPilot produces layouts with higher semantic alignment and structural coherence.}
    \label{fig:qualitative comparison}
    % \vspace{-0.5em}
\end{figure*}

%需要说明的是，参考布局并不被视为该文本描述下唯一正确或最优的设计方案，而是作为一个明确的“目标意图”用于评估可控性。在很多交互式场景中，用户通常会给出一份具体的参考布局（或草图 / IDSL 脚本），并希望系统能够尽可能贴合该方案进行生成或编辑。因此，Layout Fidelity (LF) 度量的是模型在给定目标布局条件下的几何跟随能力，而非设计空间中所有合理解的多样性。此外，我们还报告了约束满足率（CSR\textsubscript{count} / CSR\textsubscript{rel}）以及用户研究结果，以从数量约束、关系约束和人类偏好多个维度综合评价方法的质量。

In our experiments, we aim to answer the following questions:
\textbf{Q1:} How does RoomPilot perform compared with existing indoor scene generation methods in terms of controllability, visual quality, and physical plausibility?
\textbf{Q2:} Does RoomPilot effectively enhance the alignment between multi-modal inputs and generated 3D layouts?
\textbf{Q3:} How critical are the individual components of RoomPilot to the overall controllability and generation quality?

% In our experiments, we investigate three key questions:
% \textbf{Q1:} How does IDSL compare with existing indoor scene generation methods in controllability, visual quality, and physical plausibility?
% \textbf{Q2:} To what extent does IDSL improve alignment between multi-modal inputs and the generated 3D layouts?
% \textbf{Q3:} How important are the individual components of IDSL for achieving overall controllability and generation quality?

\subsection{Experimental Setup}

\paragraph{Evaluation Settings.} We quantitatively evaluate RoomPilot on open-vocabulary scene generation, following \cite{yang2025sceneweaver, sun2025layoutvlm}, which evaluates generation across diverse room type descriptions.
For each room type, we generate and evaluate 30 scenes using natural language prompts that describe room configuration, furniture types, and spatial relationships. 
See Appendix \ref{sec:details of experiments_supp} for more details.
\paragraph{Baselines.} We compare RoomPilot with six representative 3D scene synthesis method, including Holodeck \cite{yang2024holodeck}, I-Design \cite{ccelen2024design}, SceneTeller~\cite{ocal2024sceneteller}, LayoutVLM \cite{sun2025layoutvlm}, SceneWeaver \cite{yang2025sceneweaver} and HSM \cite{pun2025hsm} on the open-vocabulary setting. Since LayoutGPT~\cite{feng2024layoutgpt} and SceneTeller~\cite{ocal2024sceneteller} are inherently limited to the living room and bedroom, we modify their prompts and constraints to support the open-vocabulary setting. 
We use the default configuration and data for each baseline itself and evaluate them under identical prompt and room configurations to ensure fairness.
This setting enables a direct comparison of each method's ability to generalize beyond the room categories originally considered in prior work.
\paragraph{Evaluation Metrics.} 
For quantitative evaluation, we adopt a combination of physical, visual, and semantic metrics following \cite{yang2025sceneweaver, ccelen2024design}. 
Physical plausibility and realism are assessed by the average number of objects in the scene (\#Obj), the number of out-of-boundary objects (\#OB), and the number of collided object pairs (\#CN). 
To evaluate visual quality and semantic consistency with the user descriptions, we compute scores for visual realism (Real.), functionality (Func.), layout correctness (Lay.), and scene completeness (Comp.). 
Following previous practice, these metrics are estimated by GPT-4, which receives the user query together with top-down renderings of the generated scenes as input.
Furthermore, we employ Layout Fidelity (LF), Constraint Satisfaction Rate on object count (CSR\textsubscript{count}), and Constraint Satisfaction Rate on geometric relations (CSR\textsubscript{rel}) to comprehensively assess the controllability of generated scenes. 
Specifically, LF measures the spatial consistency between generated and reference layouts, CSR\textsubscript{count} reflects how well the object quantities adhere to the given constraints, and CSR\textsubscript{rel} assesses the correctness of geometric and functional relationships among objects.
See Appendix \ref{sec:details of experiments_supp} for more details.

\begin{table}[t]
\centering
\caption{CAD-conditioned scene generation comparison. 
% Since no existing method natively accepts CAD floor plans 
% as input, we pass CAD-derived room descriptions generated 
% by our Cross-Modal Semantic Parsing module to Holodeck as 
% text input. Geometric fidelity metrics are computed against 
% the input CAD floor plans.
}
\resizebox{\linewidth}{!}{
\begin{tabular}{lcccccccccccc}
\toprule
& \multicolumn{3}{c}{Geometric Fidelity} 
& \multicolumn{3}{c}{Physics}
& \multicolumn{4}{c}{Visual \& Semantics} \\
\cmidrule(lr){2-4} \cmidrule(lr){5-7} \cmidrule(lr){8-11}
Method 
& Wall Acc.$\uparrow$ & Area Err.$\downarrow$ & Open. Acc.$\uparrow$
& \#Obj$\uparrow$ & \#OB$\downarrow$ & \#CN$\downarrow$
& Real.$\uparrow$ & Func.$\uparrow$ & Lay.$\uparrow$ & Comp.$\uparrow$ \\
\midrule
Holodeck~\cite{yang2024holodeck} 
& 0.31 & 0.48 & 0.22 
& 11.3 & 1.2 & 0.2 
& 6.2 & 6.0 & 4.8 & 3.5 \\
\textbf{RoomPilot (CAD)}          
& \textbf{0.87} & \textbf{0.09} & \textbf{0.81} 
& \textbf{18.5} & \textbf{0.0} & \textbf{0.0}
& \textbf{8.5} & \textbf{8.2} & \textbf{7.8} & \textbf{7.6} \\
\bottomrule
\end{tabular}
}
\label{tab:cad_eval}
% \vspace{-0.5em}
\end{table}

\begin{figure}[t]
    \centering
    \includegraphics[width=\linewidth]{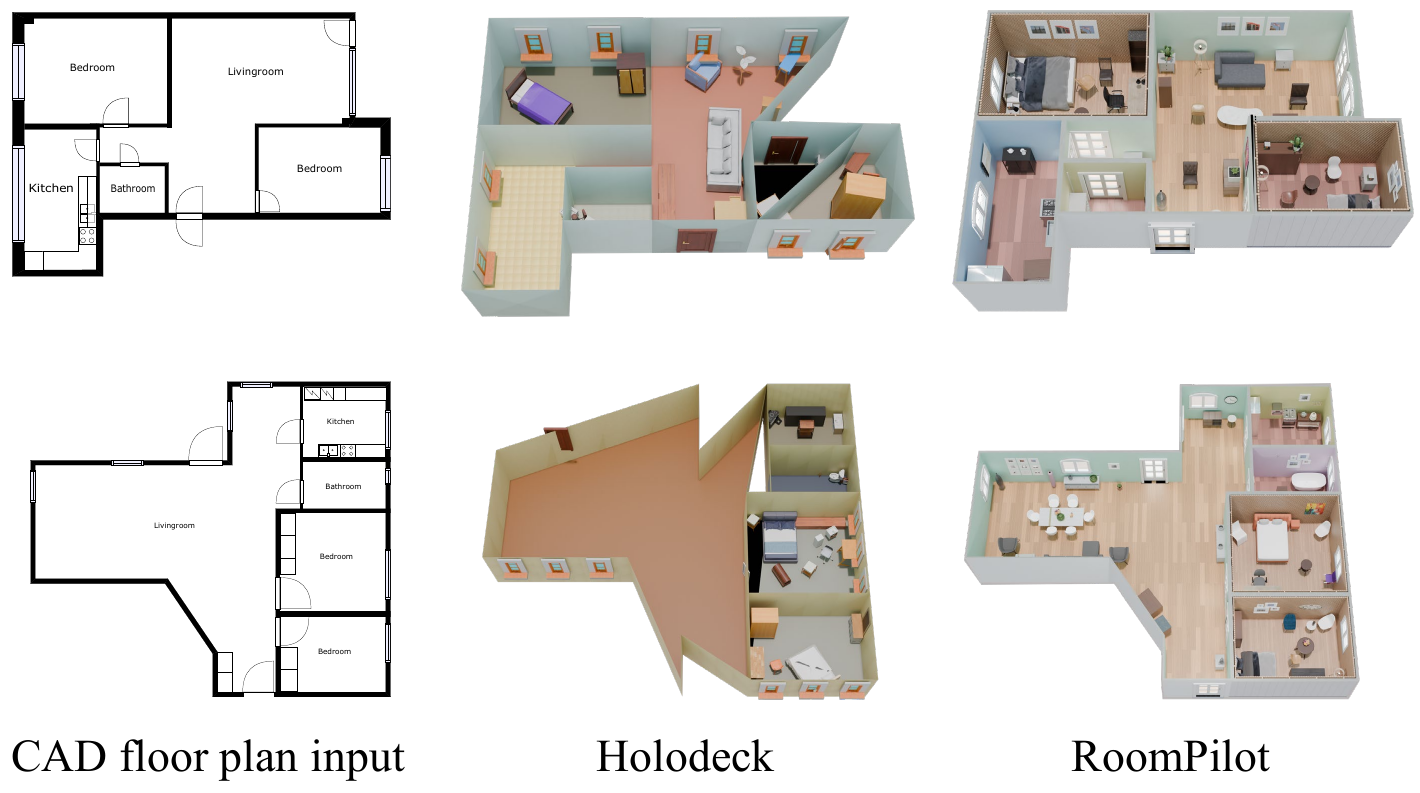}
    \caption{\textbf{CAD-conditioned scene generation comparison.}
    % RoomPilot faithfully preserves the architectural geometry of the input CAD floor plan, including room shapes, dimensions, and opening positions, while generating semantically coherent indoor scenes. Holodeck, adapted to receive CAD-derived text descriptions, fails to respect the input floor plan geometry and produces scenes with incorrect room layouts.
    }
    \vspace{-0.5em}
    \label{fig:cad_comparison}
    \vspace{-0.5em}
\end{figure}

\subsection{Open-vocabulary Scene Generation}

Table~\ref{tab:quantitative comparison} presents the quantitative comparison between RoomPilot and existing scene synthesis methods. Across all room types, RoomPilot achieves superior performance in both physical plausibility and visual–semantic quality, generating complete object configurations with zero out-of-boundary or collision cases and producing layouts that exhibit stronger realism, functionality, and alignment with user descriptions.
Figure~\ref{fig:qualitative comparison} provides qualitative examples across five room categories. 
% Compared with Holodeck \cite{yang2024holodeck}, I-Design \cite{ccelen2024design}, LayoutGPT \cite{feng2024layoutgpt}, SceneTeller \cite{ocal2024sceneteller}, LayoutVLM \cite{sun2024layoutvlm} and Scene Weaver. 
Prior methods often yield incomplete or poorly structured layouts, whereas RoomPilot consistently produces coherent spatial organization, correct object relations, and faithful adherence to textual intent.
These results show that combining IDSL with self-regulating optimization yields indoor scenes that are more structurally coherent, semantically aligned, and practically usable.
Further quantitative comparisons are provided in Appendix \ref{sec:additional experimental results_supp}.

\begin{table}[t]
\centering
\caption{Comparison of controllability performance across recent scene synthesis methods.
% Three metrics are reported: 
% Layout Fidelity (LF, IoU-based spatial alignment), 
% Constraint Satisfaction Rate (CSR\textsubscript{count}, object quantity
}
\vspace{-0.3em}
\resizebox{0.7\linewidth}{!}{
\begin{tabular}{lccc}
\toprule
Method & 
\textbf{LF} $\uparrow$ & 
\textbf{CSR\textsubscript{count}} $\uparrow$ & 
\textbf{CSR\textsubscript{rel}} $\uparrow$ \\
\midrule
LayoutGPT~\cite{feng2024layoutgpt} & 0.05 & 0.55 & 0.12 \\
Holodeck~\cite{yang2024holodeck} & 0.23 & 0.74 & 0.42 \\
I-Design~\cite{ccelen2024design} & 0.20 & 0.69 & 0.19 \\
SceneTeller~\cite{ocal2024sceneteller} & 0.18 & 0.71 & 0.35 \\
LayoutVLM~\cite{sun2025layoutvlm} & 0.25 & 0.78 & 0.49 \\
SceneWeaver~\cite{yang2025sceneweaver} & 0.35 & 0.79 & 0.58 \\
HSM~\cite{pun2025hsm} & 0.32 & 0.75 & 0.68 \\
\midrule
\textbf{RoomPilot} & \textbf{0.58} & \textbf{0.89} & \textbf{0.79} \\
\bottomrule
\end{tabular}
}
\label{tab:controllability comparison}
\vspace{-8pt}
\end{table}

\subsection{CAD-based Scene Generation}
Since no existing method accepts CAD floor plans as input, we adapt Holodeck~\cite{yang2024holodeck} by providing room-level information extracted by our CAD parsing pipeline (Section~\ref{subsubsec:CAD Floor Plan Parsing}), including room types, dimensions, and adjacency.
We evaluate geometric fidelity via Wall Accuracy, Room Area Error, and Opening Accuracy against input CAD floor plans, along with visual-semantic metrics from Section~\ref{sec:Experiments} (see Appendix~\ref{appendix:cad_metrics} for details).
As shown in Table~\ref{tab:cad_eval} and Figure~\ref{fig:cad_comparison}, RoomPilot significantly outperforms Holodeck on all metrics. RoomPilot directly ingests CAD geometry as hard constraints, preserving room shapes, dimensions, and opening positions. Holodeck, without native support for precise geometric inputs, produces layouts that deviate from the input floor plan. 
% RoomPilot also achieves higher visual-semantic scores, indicating that precise geometric grounding improves downstream furniture placement and overall scene coherence.

\subsection{Controllability Evaluation}

We quantitatively evaluate the controllability of scene synthesis methods using three metrics: Layout Fidelity (LF), object-count constraint satisfaction (CSR\textsubscript{count}), and relational constraint satisfaction (CSR\textsubscript{rel}), see Appendix~\ref{sec:details of experiments_supp} for details. As shown in Table~\ref{tab:controllability comparison}, RoomPilot significantly outperforms existing methods across all three dimensions. Compared to diffusion- or retrieval-based approaches (e.g., LayoutGPT \cite{feng2024layoutgpt}), RoomPilot achieves higher spatial alignment and constraint adherence, indicating stronger control over geometric structure and semantic relations. Results highlight the effectiveness of IDSL and the constraint optimization framework for controllable, semantically consistent scene synthesis.

\begin{table}[t]
\centering
\caption{Ablation study of RoomPilot. We report Layout Fidelity (LF), Constraint Satisfaction Rate on object count (CSR\textsubscript{count}), and geometric/relational constraints (CSR\textsubscript{rel}).}
\resizebox{0.85\linewidth}{!}{
\begin{tabular}{lccc}
\toprule
Configuration & LF $\uparrow$ & CSR\textsubscript{count} $\uparrow$ & CSR\textsubscript{rel} $\uparrow$ \\
\midrule
\multicolumn{4}{l}{\textit{Ablating core modules}} \\
\quad \textit{w/o} IDSL & 0.42 & 0.68 & 0.47 \\
\quad \textit{w/o} Scene Optim. & 0.44 & 0.61 & 0.63 \\
\midrule
\multicolumn{4}{l}{\textit{Ablating optimization components}} \\
\quad \textit{w/o} Structural Energy & 0.50 & 0.84 & 0.78 \\
\quad \textit{w/o} Semantic Energy & 0.52 & 0.85 & 0.80 \\
\quad \textit{w/o} Adaptive Annealing & 0.54 & 0.86 & 0.84 \\
\quad \textit{w/o} Prog. Rule Activation & 0.51 & 0.83 & 0.79 \\
\midrule
\multicolumn{4}{l}{\textit{Ablating generation components}} \\
\quad \textit{w/o} PCG\textsubscript{wall} & 0.53 & 0.82 & 0.81 \\
\quad \textit{w/o} Style Compat. $\psi(\cdot)$ & 0.55 & 0.87 & 0.85 \\
\quad \textit{w/o} PCG Generators & 0.48 & 0.74 & 0.72 \\
\midrule
Full Model & \textbf{0.58} & \textbf{0.89} & \textbf{0.90} \\
\bottomrule
\end{tabular}
}
\label{tab:ablation_study}
% \vspace{-1.3em}
\end{table}

\subsection{Ablation Study}

\begin{table}[tb]
\centering
\caption{User study results.}
\label{tab:user study}
\vspace{-0.3em}
\resizebox{0.8\linewidth}{!}{
\begin{tabular}{lccc}
\toprule
Method & \makecell{Visual \\ Quality $\uparrow$} & \makecell{Layout \\ Reasonableness $\uparrow$} & \makecell{Physical \\ Plausibility $\uparrow$} \\
\midrule
LayoutGPT~\cite{feng2024layoutgpt} & 3.35 & 3.27 & 3.45 \\
Holodeck~\cite{yang2024holodeck}  & 3.09 & 3.02 & 3.14 \\
I-Design~\cite{ccelen2024design} & 3.23& 3.10 & 3.19  \\
SceneTeller~\cite{ocal2024sceneteller} & 3.27 & 3.32 & 3.43  \\
Infinigen~\cite{raistrick2024infinigen} & 3.35 & 3.23 & 3.35 \\
LayoutVLM~\cite{sun2025layoutvlm} & 3.26 & 3.10& 3.22  \\
Sceneweaver~\cite{yang2025sceneweaver} & 3.42 & 3.31& 3.43  \\
HSM~\cite{pun2025hsm} & 3.68 & 3.87 & 4.02 \\
\midrule
\textbf{RoomPilot (Ours)} & \textbf{4.10} & \textbf{4.31} & \textbf{4.42} \\
\bottomrule
\end{tabular}
}
\vspace{-0.8em}
\end{table}

As shown in Table~\ref{tab:ablation_study}, each component of RoomPilot contributes to final performance. Removing IDSL causes the largest drop, confirming structured intent specification is critical for controllable layout generation. Disabling scene optimization mainly reduces relational and geometric consistency, while ablating individual energies or rule activation leads to moderate, consistent degradation. On the generation side, removing PCG generators causes a clear drop, whereas wall-aware generation and style compatibility provide complementary gains. Overall, the full model achieves the best balance between layout fidelity and constraint satisfaction.
The runtime analysis of RoomPilot is given in Appendix~\ref{sec:runtime}.

\subsection{Human Study}

To further assess the quality of the generated scenes, we conducted a user study with 30 participants, where each participant evaluated 10 scenes from each method. The evaluation metrics and results are summarized in Table~\ref{tab:user study}. The results show that RoomPilot consistently achieves the highest scores across all dimensions, demonstrating superior visual quality, layout reasonableness, and physical plausibility compared with existing approaches.

\section{Conclusion}
\label{sec:Conclusion}
% We present RoomPilot, a unified framework for controllable, high-fidelity indoor scene synthesis from textual descriptions or CAD floor plans. At its core, the Indoor Domain-Specific Language (IDSL) provides a structured semantic--geometric representation for coherent multimodal parsing and interpretable design specification. Combined with a self-regulating optimization mechanism and a hierarchical generation pipeline, RoomPilot produces multi-room scenes that are structurally consistent, semantically aligned, and physically plausible. Extensive quantitative experiments, ablations, and user studies show clear improvements over existing approaches in controllability, visual quality, and usability.

We present RoomPilot, a unified framework for controllable, high-fidelity indoor scene synthesis from text descriptions or CAD floor plans. Its core, the Indoor Domain-Specific Language (IDSL), provides a structured semantic--geometric representation for coherent multimodal parsing and interpretable design specification. Combined with self-regulating optimization and hierarchical generation, RoomPilot produces structurally consistent, semantically aligned, and physically plausible multi-room scenes, outperforming prior methods in controllability, visual quality, and usability.

\bibliographystyle{ACM-Reference-Format}
\bibliography{sample-base}

%%
%% If your work has an appendix, this is the place to put it.
\newpage
\appendix

\section{Details of Cross-Modal Semantic Parsing}
\label{sec:parsing_supp}

\subsection{Text Description Parsing Details}

The text description parsing processes free-form textual descriptions to extract semantic entities (e.g., rooms, objects, attributes) and their spatial and functional relations. The output of this stage is a structured IDSL that captures both the semantic intent and the geometric constraints described in the text.

Given a textual scene description \( T \), we employ GPT-4o to extract the underlying scene elements. The LLM parsing process is formulated as a composite mapping \( \Psi \circ \Phi: T \rightarrow \mathcal{S}_0 \), converting free-form language into a structured representation compatible with IDSL.

\paragraph{Entity Extraction.} The lexical–semantic stage \( \Phi_1: T \rightarrow E \) identifies spatial entities \( E = \{\text{objects},\ \text{rooms},\ \text{attributes}\} \) and their semantic annotations. These include identifying the rooms (e.g., living room, kitchen), objects (e.g., sofa, table), and attributes such as size, position, and orientation.

\paragraph{Relation Extraction.} In the relation extraction stage \( \Phi_2: T \rightarrow R \), spatial, functional, and hierarchical relations $R=$ \{\text{spatial},\ \text{functional},\ \text{hierarchical}\}  are recovered, capturing positional dependencies and functional intent. For example, relations such as "sofa is next to the wall" or "bed is placed in the corner" are identified and structured.

\paragraph{Semantic Structuring.} The semantic structuring stage \( \Phi_3: (E, R) \rightarrow G \) organizes the extracted entities and relations into a semantic graph \( G = (V, E_G) \), where nodes correspond to entities (rooms, objects), and edges encode the identified relations. This graph provides a high-level view of the scene, with connections indicating spatial and functional dependencies.

\paragraph{IDSL Mapping.} Finally, the mapping \( \Psi \) converts this semantic graph into an initial IDSL \( \mathcal{S}_0 \), where symbolic rules and parametric preferences are normalized into the forms required by the downstream optimizer. For example, the relationships between rooms and objects are translated into shared edges or spatial constraints, while the rooms are assigned appropriate room polygons and objects are placed based on their identified relationships.

The resulting IDSL includes both symbolic and parametric data. Rooms are represented by their contours and labeled accordingly, while objects are described with their positions, sizes, and orientations, all encoded in a format compatible with optimization phase.

\subsection{Prompt Design for Text-to-IDSL Parsing}
\label{app:prompt-design}

We provide the complete prompt templates used in our hierarchical text-to-IDSL parsing pipeline (Section~\ref{subsubsec:Text Description Parsing}). The parsing is decomposed into three stages aligned with the three IDSL levels: building-level extracts global structure, room-level plans furniture configurations, and object-level determines spatial arrangements. Relation constraints are explicitly defined in the prompt schemas and instantiated through structured \texttt{relation\_graph} entries, as illustrated in the few-shot examples below.

\subsubsection{Building-Level Prompt}
\label{app:prompt-building}

The building-level prompt extracts global scene structure including all room contours and structural openings (doors, windows, open passages), ensuring geometric consistency within a single LLM context.

\begin{lstlisting}[language=json, style=idslstyle, caption={Building-level prompt template.}, label={lst:prompt-building}]
[System Message]
You are an indoor scene parser. Given a textual
description of an indoor environment, extract the
building-level structure including room layout
geometry and structural openings. Output a JSON
object following the IDSL building-level schema below.

## IDSL Building-Level Schema
{
  "building": {
    "building_id": <string>,
    "scene_tags": ["Semantics(building)"],
    "room_entities": [<room_id>, ...],
    "relations": [{
      "neighbours": [[<int>, ...], ...],
      "rooms": [<room_id>, ...],
      "entrance": <int>
    }]
  },
  "rooms": {
    "<room_id>": {
      "room_type": <string>,
      "room_contour": [[x, y], ...],
      "area": <float>,
      "bounds": [x_min, y_min, x_max, y_max],
      "room_dimensions": {
        "width": <float>,
        "length": <float>,
        "height": 3.0
      },
      "room_tags": [
        "Semantics(room)", "Semantics(<type>)"
      ],
      "room_relations": [{
        "relation_type": "SharedEdge",
        "target_room": <adjacent_room_id>
      }, ...]
    }
  },
  "openings": {
    "<opening_id>": {
      "semantic_tags": [
        "Semantics(cutter)", "Semantics(<type>)"
      ],
      "position_world": [x, y, z],
      "relation_graph": [{
        "relation": {"relation_type": "CutFrom"},
        "target_name": <room_id>
      }, ...]
    }
  },
  "room_segments": {
    "<room_id>": <text fragment for this room>
  }
}

## Field Descriptions
- "room_entities": list of room identifiers, formatted
  as "<room_type>_0/<index>" (e.g., "bedroom_0/0").
- "relations.neighbours": adjacency list where each
  entry lists indices of neighboring rooms.
- "relations.rooms": ordered room list corresponding
  to the adjacency list; include "exterior_0/0" as the
  last entry for rooms with exterior walls.
- "relations.entrance": index of the room whose door
  has a CutFrom relation to exterior_0/0.
- "rooms": per-room geometry. room_contour is a closed
  2D polygon; adjacent rooms MUST share identical edge
  coordinates where they have a SharedEdge relation.
- "openings": doors, windows, and open passages that
  connect rooms or face the exterior.
  - Doors: Semantics(door), connect two rooms.
  - Windows: Semantics(window), typically on exterior
    walls.
  - Open passages: Semantics(open), represent archways
    or open connections between rooms without a door.
  Each opening has a CutFrom relation to the room(s)
  it is cut from. Doors and open passages have two
  CutFrom entries (one per adjacent room); windows
  facing the exterior have one CutFrom entry.
- "room_segments": map each room_id to the substring
  of the user description that pertains to that room.
  If a sentence describes relations across multiple
  rooms, include it in all involved rooms' segments.

## Rules
1. Infer room types from context (e.g., "three-bedroom"
   implies three bedroom rooms).
2. Infer reasonable adjacency from common floor plan
   conventions if not explicitly stated.
3. Generate all room contours jointly to ensure
   geometric consistency: adjacent rooms must share
   exact edge coordinates, and room contours must tile
   without gaps or overlaps.
4. Estimate room dimensions based on room type and
   typical residential proportions. Soft preferences
   (e.g., "as large as possible") should be reflected
   in the generated geometry.
5. Do NOT generate furniture or decorative objects.
   Only generate structural openings (doors, windows,
   open passages) that connect rooms or face the
   exterior.
6. Place doors on shared edges between adjacent rooms.
   Place windows on exterior walls based on room type
   conventions (e.g., bedrooms and living rooms
   typically have large windows; bathrooms may have
   small or no windows).
7. Output valid JSON only, no commentary.

## Example
Input: "A two-bedroom apartment with a living room
and kitchen. The kitchen is next to the living room
with an open passage between them."

Output:
{
  "building": {
    "building_id": "suite_example",
    "scene_tags": ["Semantics(building)"],
    "room_entities": [
      "living-room_0/0", "kitchen_0/0",
      "bedroom_0/0", "bedroom_0/1"
    ],
    "relations": [{
      "neighbours": [[1,4],[0,2],[1],[3],[0]],
      "rooms": [
        "living-room_0/0", "kitchen_0/0",
        "bedroom_0/0", "bedroom_0/1",
        "exterior_0/0"
      ],
      "entrance": 0
    }]
  },
  "rooms": {
    "living-room_0/0": {
      "room_type": "living-room",
      "room_contour": [
        [0.0,0.0],[5.0,0.0],[5.0,4.0],
        [0.0,4.0],[0.0,0.0]
      ],
      "area": 20.0,
      "bounds": [0.0,0.0,5.0,4.0],
      "room_dimensions": {
        "width":5.0, "length":4.0, "height":3.0
      },
      "room_tags": ["Semantics(room)",
        "Semantics(living-room)"],
      "room_relations": [{
        "relation_type": "SharedEdge",
        "target_room": "kitchen_0/0"
      }]
    },
    "kitchen_0/0": {
      "room_type": "kitchen",
      "room_contour": [
        [5.0,0.0],[8.0,0.0],[8.0,4.0],
        [5.0,4.0],[5.0,0.0]
      ],
      "area": 12.0,
      "bounds": [5.0,0.0,8.0,4.0],
      "room_dimensions": {
        "width":3.0, "length":4.0, "height":3.0
      },
      "room_tags": ["Semantics(room)",
        "Semantics(kitchen)"],
      "room_relations": [{
        "relation_type": "SharedEdge",
        "target_room": "living-room_0/0"
      },{
        "relation_type": "SharedEdge",
        "target_room": "bedroom_0/0"
      }]
    },
    "bedroom_0/0": {
      "room_type": "bedroom",
      "room_contour": [
        [5.0,4.0],[8.0,4.0],[8.0,8.0],
        [5.0,8.0],[5.0,4.0]
      ],
      "area": 12.0,
      "bounds": [5.0,4.0,8.0,8.0],
      "room_dimensions": {
        "width":3.0, "length":4.0, "height":3.0
      },
      "room_tags": ["Semantics(room)",
        "Semantics(bedroom)"],
      "room_relations": [{
        "relation_type": "SharedEdge",
        "target_room": "kitchen_0/0"
      }]
    },
    "bedroom_0/1": {
      "room_type": "bedroom",
      "room_contour": [
        [0.0,4.0],[5.0,4.0],[5.0,8.0],
        [0.0,8.0],[0.0,4.0]
      ],
      "area": 20.0,
      "bounds": [0.0,4.0,5.0,8.0],
      "room_dimensions": {
        "width":5.0, "length":4.0, "height":3.0
      },
      "room_tags": ["Semantics(room)",
        "Semantics(bedroom)"],
      "room_relations": []
    }
  },
  "openings": {
    "door_0": {
      "semantic_tags": [
        "Semantics(cutter)", "Semantics(door)"
      ],
      "position_world": [2.5, 0.0, 1.1],
      "relation_graph": [{
        "relation": {"relation_type": "CutFrom"},
        "target_name": "living-room_0/0"
      },{
        "relation": {"relation_type": "CutFrom"},
        "target_name": "exterior_0/0"
      }]
    },
    "open_0": {
      "semantic_tags": [
        "Semantics(cutter)", "Semantics(open)"
      ],
      "position_world": [5.0, 2.0, 1.1],
      "relation_graph": [{
        "relation": {"relation_type": "CutFrom"},
        "target_name": "living-room_0/0"
      },{
        "relation": {"relation_type": "CutFrom"},
        "target_name": "kitchen_0/0"
      }]
    },
    "window_0": {
      "semantic_tags": [
        "Semantics(cutter)", "Semantics(window)"
      ],
      "position_world": [6.5, 0.0, 1.8],
      "relation_graph": [{
        "relation": {"relation_type": "CutFrom"},
        "target_name": "kitchen_0/0"
      }]
    }
  },
  "room_segments": {
    "living-room_0/0": "a living room",
    "kitchen_0/0": "kitchen, next to the living room
      with an open passage between them",
    "bedroom_0/0": "",
    "bedroom_0/1": ""
  }
}

[User Message]
<user's full scene description>
\end{lstlisting}

Note that in the example above, \texttt{living-room\_0/0} and \texttt{kitchen\_ \\ 0/0} share the edge at $x=5.0$, demonstrating geometric consistency. The \texttt{open\_0} entry shows how an open passage is represented with two \texttt{CutFrom} relations, while \texttt{window\_0} has a single \texttt{CutFrom} to its room (exterior windows do not reference \texttt{exterior\_0/0} in the opening's relation graph, as the exterior association is implicit from the window's position on an exterior wall). The entrance is identified as room index 0 (\texttt{living-room\_0/0}), whose \texttt{door\_0} connects to \texttt{exterior\_0/0}.

\subsubsection{Room-Level Prompt}
\label{app:prompt-room}

The room-level prompt is invoked once per room. It receives the building-level context and room-specific description, and generates a furniture plan: what objects to place, their categories, and functional roles. This stage determines the \emph{composition} of each room without specifying precise spatial arrangements.

\begin{lstlisting}[language=json, style=idslstyle, caption={Room-level prompt template.}, label={lst:prompt-room}]
[System Message]
You are an indoor room planner. Given a room's fixed
geometry and its description, generate a furniture
plan listing all objects to be placed in this room.
For each object, specify its category, semantic role,
and any spatial preferences mentioned in the
description. Do NOT generate precise positions or
bounding boxes at this stage.
Output a JSON object following the schema below.

## Building-Level Context (from previous step)
<inserted building-level JSON output>

## Current Room (geometry is fixed)
Room ID: <room_id>
Room Type: <room_type>
Room Contour: <room_contour from building-level>
Bounds: <bounds from building-level>
Room Dimensions: <room_dimensions from building-level>
Adjacent Rooms: [<list of adjacent room_ids>]
Openings in this room: [<list of openings with
  positions from building-level>]
Room Description: <text fragment from room_segments>

## Furniture Plan Schema
{
  "furniture_plan": [{
    "object_name": <string, Category_id format>,
    "category": <string>,
    "semantic_tags": ["Semantics(<tag>)", ...],
    "placement_preference": <string or null>,
    "wall_adjacent": <bool>
  }, ...]
}

## Field Descriptions
- "object_name": unique identifier in Category_id
  format (e.g., "DiningTable_001", "Chair_001").
- "category": furniture category (e.g., "diningtable",
  "chair", "bed", "sofa", "largeshelf").
- "semantic_tags": semantic labels describing the
  object's function (e.g., Semantics(furniture),
  Semantics(seating), Semantics(storage)).
- "placement_preference": spatial preference from the
  user description if any (e.g., "in the center",
  "against the wall", "near the window", "facing the
  TV"). Set to null if no preference is specified.
- "wall_adjacent": whether this object should be
  placed against a wall (true for beds, shelves,
  desks, cabinets, TV stands; false for tables,
  chairs in the center).

## Rules
1. Include all objects mentioned in the description.
2. If the description is vague or empty, infer typical
   furniture based on room type (e.g., a bedroom
   typically contains a bed, nightstands, a wardrobe).
3. Respect object counts specified in the description
   (e.g., "four chairs" -> 4 chair entries).
4. Extract placement preferences verbatim from the
   description when available.
5. Output valid JSON only, no commentary.

## Example
Room: "dining-room_0/0" (6.0 x 6.0 m)
Description: "a dining room with a dining table in
  the center, four chairs around the table, and a
  large shelf against the wall near the window"

Output:
{
  "furniture_plan": [
    {
      "object_name": "DiningTable_001",
      "category": "diningtable",
      "semantic_tags": ["Semantics(furniture)",
        "Semantics(table)"],
      "placement_preference": "in the center",
      "wall_adjacent": false
    },
    {
      "object_name": "Chair_001",
      "category": "chair",
      "semantic_tags": ["Semantics(seating)",
        "Semantics(furniture)"],
      "placement_preference": "around the table",
      "wall_adjacent": false
    },
    {
      "object_name": "Chair_002",
      "category": "chair",
      "semantic_tags": ["Semantics(seating)",
        "Semantics(furniture)"],
      "placement_preference": "around the table",
      "wall_adjacent": false
    },
    {
      "object_name": "Chair_003",
      "category": "chair",
      "semantic_tags": ["Semantics(seating)",
        "Semantics(furniture)"],
      "placement_preference": "around the table",
      "wall_adjacent": false
    },
    {
      "object_name": "Chair_004",
      "category": "chair",
      "semantic_tags": ["Semantics(seating)",
        "Semantics(furniture)"],
      "placement_preference": "around the table",
      "wall_adjacent": false
    },
    {
      "object_name": "LargeShelf_001",
      "category": "largeshelf",
      "semantic_tags": ["Semantics(furniture)",
        "Semantics(storage)"],
      "placement_preference":
        "against the wall near the window",
      "wall_adjacent": true
    }
  ]
}

[User Message]
Room description for <room_id>:
<room-specific text fragment from Step 1>
\end{lstlisting}

\subsubsection{Object-Level Prompt}
\label{app:prompt-object}

The object-level prompt is invoked once per room. It receives the building-level context, the room-level furniture plan, and the room-specific description. It generates precise positions, bounding boxes, orientations, and \texttt{relation\_graph} entries for each object in the furniture plan.

\begin{lstlisting}[language=json, style=idslstyle, caption={Object-level prompt template.}, label={lst:prompt-object}]
[System Message]
You are an indoor object placement engine. Given a
room's fixed geometry, its openings, and a furniture
plan, generate the spatial configuration for each
object. The room contour, dimensions, and openings
are already determined -- do NOT modify them. The
furniture list is already determined -- do NOT add
or remove objects. Only generate positions, bounding
boxes, orientations, and relation_graph entries.
Output a JSON object following the schema below.

## Building-Level Context (from Step 1)
<inserted building-level JSON output>

## Current Room (geometry is fixed)
Room ID: <room_id>
Room Type: <room_type>
Room Contour: <room_contour from building-level>
Bounds: <bounds from building-level>
Openings in this room: [<list of openings with
  positions from building-level>]
Room Description: <text fragment from room_segments>

## Furniture Plan (from Step 2)
<inserted furniture plan JSON output>

## IDSL Object-Level Schema
{
  "<object_name>": {
    "category": <string>,
    "semantic_tags": ["Semantics(<tag>)", ...],
    "position_world": [x, y, z],
    "rotation_euler": [rx, ry, rz],
    "bounding_box": {
      "min": [x,y,z], "max": [x,y,z],
      "center": [x,y,z], "size": [sx,sy,sz]
    },
    "relation_graph": [{
      "relation": {
        "relation_type": <string>,
        "margin": <float>,
        "check_z": <bool>,
        "child_tags": [<string>, ...],
        "parent_tags": [<string>, ...]
      },
      "target_name": <room_id or object_name>,
      "child_plane_idx": <int>,
      "parent_plane_idx": <int>
    }, ...]
  }
}

## Relation Types and Instantiation

"StableAgainst" encodes physical support and contact
between an object surface and a target surface. It is
instantiated through child_tags (which surface of the
object) and parent_tags (which surface of the target):

  Floor support (object rests on floor):
    child_tags: ["Subpart(bottom)"]
    parent_tags: ["Subpart(support)"]
    -> object's bottom surface on room's floor

  Wall leaning (object back against wall):
    child_tags: ["Subpart(back)"]
    parent_tags: ["Subpart(wall)"]
    margin: 0.07 (small gap from wall)
    -> object's back surface against room's wall

  Surface placement (object on top of furniture):
    child_tags: ["Subpart(bottom)"]
    parent_tags: ["Subpart(top)"]
    -> object's bottom on another object's top surface

Every object MUST have at least one "StableAgainst"
relation for floor support. Objects with
wall_adjacent=true in the furniture plan should
additionally have a wall-leaning relation.

"CutFrom" is used only for structural openings and
is already generated in the building-level pass.
Do NOT generate CutFrom relations at this stage.

## Rules
1. ALL objects must be placed strictly within the
   room's bounds.
2. Do NOT place objects that block doors, windows, or
   open passages. Maintain clearance around all
   openings listed in the building-level context.
3. Respect placement_preference from the furniture
   plan (e.g., "in the center" -> position near room
   center; "against the wall" -> position near wall
   with wall-leaning relation).
4. Assign at least one "StableAgainst" relation per
   object (floor support).
5. Objects with wall_adjacent=true should have a
   second "StableAgainst" with Subpart(wall).
6. Estimate reasonable bounding boxes based on common
   furniture sizes. These serve as initial estimates
   and will be refined by downstream optimization.
7. Output valid JSON only, no commentary.

## Example
Room: "dining-room_0/0" (bounds: [1.5,1.5,7.5,7.5])
Description: "a dining room with a dining table in
  the center, four chairs around the table, and a
  large shelf against the wall near the window"
Openings: [door_0 at (7.5, 5.5, 1.1),
  window_0 at (4.1, 1.5, 1.8)]
Furniture Plan:
  DiningTable_001 (diningtable, center)
  Chair_001..004 (chair, around the table)
  LargeShelf_001 (largeshelf, against wall near window)

Output:
{
  "objs": {
    "DiningTable_001": {
      "category": "diningtable",
      "semantic_tags": ["Semantics(furniture)",
        "Semantics(table)", "Semantics(object)"],
      "position_world": [4.5, 4.5, 0.114],
      "rotation_euler": [0.0, 0.0, 0.0],
      "bounding_box": {
        "min": [3.5,3.5,0.114],
        "max": [5.5,5.5,0.874],
        "center": [4.5,4.5,0.494],
        "size": [2.0,2.0,0.760]
      },
      "relation_graph": [{
        "relation": {
          "relation_type": "StableAgainst",
          "margin": 0.0, "check_z": true,
          "child_tags": ["Subpart(bottom)"],
          "parent_tags": ["Subpart(support)"]
        },
        "target_name": "dining-room_0/0",
        "child_plane_idx": 0,
        "parent_plane_idx": 0
      }]
    },
    "Chair_001": {
      "category": "chair",
      "semantic_tags": ["Semantics(seating)",
        "Semantics(furniture)", "Semantics(chair)"],
      "position_world": [3.134, 3.641, 0.573],
      "rotation_euler": [0.0, 0.0, -2.356],
      "bounding_box": {
        "min": [2.548,3.055,0.114],
        "max": [3.427,3.934,1.028],
        "center": [2.988,3.495,0.571],
        "size": [0.879,0.879,0.914]
      },
      "relation_graph": [{
        "relation": {
          "relation_type": "StableAgainst",
          "margin": 0.0, "check_z": true,
          "child_tags": ["Subpart(bottom)"],
          "parent_tags": ["Subpart(support)"]
        },
        "target_name": "dining-room_0/0",
        "child_plane_idx": 0,
        "parent_plane_idx": 1
      }]
    },
    "Chair_002": {
      "category": "chair",
      "semantic_tags": ["Semantics(seating)",
        "Semantics(furniture)", "Semantics(chair)"],
      "position_world": [5.785, 5.763, 0.573],
      "rotation_euler": [0.0, 0.0, 1.571],
      "bounding_box": {
        "min": [5.449,5.684,0.114],
        "max": [6.121,6.254,1.028],
        "center": [5.785,5.969,0.571],
        "size": [0.672,0.570,0.914]
      },
      "relation_graph": [{
        "relation": {
          "relation_type": "StableAgainst",
          "margin": 0.0, "check_z": true,
          "child_tags": ["Subpart(bottom)"],
          "parent_tags": ["Subpart(support)"]
        },
        "target_name": "dining-room_0/0",
        "child_plane_idx": 0,
        "parent_plane_idx": 4
      }]
    },
    "Chair_003": {
      "category": "chair",
      "semantic_tags": ["Semantics(seating)",
        "Semantics(furniture)", "Semantics(chair)"],
      "position_world": [6.956, 6.489, 0.573],
      "rotation_euler": [0.0, 0.0, -1.571],
      "bounding_box": {
        "min": [6.619,5.998,0.114],
        "max": [7.292,6.568,1.026],
        "center": [6.956,6.283,0.570],
        "size": [0.673,0.570,0.912]
      },
      "relation_graph": [{
        "relation": {
          "relation_type": "StableAgainst",
          "margin": 0.0, "check_z": true,
          "child_tags": ["Subpart(bottom)"],
          "parent_tags": ["Subpart(support)"]
        },
        "target_name": "dining-room_0/0",
        "child_plane_idx": 0,
        "parent_plane_idx": 1
      }]
    },
    "Chair_004": {
      "category": "chair",
      "semantic_tags": ["Semantics(seating)",
        "Semantics(furniture)", "Semantics(chair)"],
      "position_world": [5.796, 2.029, 0.573],
      "rotation_euler": [0.0, 0.0, -3.142],
      "bounding_box": {
        "min": [5.305,1.692,0.114],
        "max": [5.875,2.365,1.029],
        "center": [5.590,2.029,0.572],
        "size": [0.570,0.673,0.915]
      },
      "relation_graph": [{
        "relation": {
          "relation_type": "StableAgainst",
          "margin": 0.0, "check_z": true,
          "child_tags": ["Subpart(bottom)"],
          "parent_tags": ["Subpart(support)"]
        },
        "target_name": "dining-room_0/0",
        "child_plane_idx": 0,
        "parent_plane_idx": 1
      }]
    },
    "LargeShelf_001": {
      "category": "largeshelf",
      "semantic_tags": ["Semantics(furniture)",
        "Semantics(storage)", "Semantics(object)"],
      "position_world": [6.652, 1.829, 0.114],
      "rotation_euler": [0.0, 0.0, 1.571],
      "bounding_box": {
        "min": [6.024,1.684,0.114],
        "max": [7.279,1.967,1.977],
        "center": [6.652,1.826,1.046],
        "size": [1.255,0.283,1.863]
      },
      "relation_graph": [{
        "relation": {
          "relation_type": "StableAgainst",
          "margin": 0.0, "check_z": true,
          "child_tags": ["Subpart(bottom)"],
          "parent_tags": ["Subpart(support)"]
        },
        "target_name": "dining-room_0/0",
        "child_plane_idx": 0,
        "parent_plane_idx": 3
      },{
        "relation": {
          "relation_type": "StableAgainst",
          "margin": 0.07, "check_z": true,
          "child_tags": ["Subpart(back)"],
          "parent_tags": ["Subpart(wall)"]
        },
        "target_name": "dining-room_0/0",
        "child_plane_idx": 0,
        "parent_plane_idx": 2
      }]
    }
  }
}

[User Message]
Room description for <room_id>:
<room-specific text fragment from Step 1>
Furniture plan:
<furniture plan JSON from Step 2>
\end{lstlisting}

The \texttt{Relation Types and Instantiation} section above directly addresses how abstract spatial descriptions (``in the center'', ``around the table'', ``against the wall near the window'') are mapped to concrete IDSL \texttt{relation\_graph} entries with specific \texttt{child\_tags}/\texttt{parent\_tags} combinations. For instance, ``a large shelf against the wall'' is instantiated as two \texttt{StableAgainst} relations: one for floor support (\texttt{Subpart(bottom)} on \texttt{Subpart(support)}) and one for wall contact (\texttt{Subpart(back)} on \texttt{Subpart(wall)} with margin 0.07). The \texttt{placement\_preference} from the furniture plan guides the object-level prompt in determining appropriate positions and orientations.

\subsubsection{Post-Assembly Validation}
\label{app:validator}

After assembling the building-level structure (rooms, openings), room-level furniture plans, and all per-room object configurations into a complete IDSL $\mathcal{S}_0$, a rule-based validator performs the following checks:
\begin{enumerate}
    \item \textit{Schema compliance}: all required fields are present and correctly typed.
    \item \textit{Topological consistency}: every \texttt{room\_id} referenced in object \texttt{relation\_graph} entries exists in the building-level \texttt{room\_entities}, and \texttt{SharedEdge} relations between rooms are symmetric (if room A declares a shared edge with room B, room B must reciprocate).
    \item \textit{Boundary consistency}: all object bounding boxes fall within their assigned room's \texttt{bounds}.
    \item \textit{Plan consistency}: every object in the furniture plan has a corresponding entry in the object-level output, and no extra objects are introduced.
\end{enumerate}
Violations of types (1), (2), and (4) trigger re-invocation of GPT-4 with the specific error message (up to 3 retries). Violations of type (3) are retained as soft constraints and resolved by the downstream self-regulating optimization (Section~\ref{subsec:optimization}), which iteratively refines object positions to achieve physically plausible layouts. The building \texttt{floor\_outline} is computed as the outer boundary of the union of all room contours during assembly.

\subsection{Text-to-IDSL Parsing Example}
\label{app:parsing-example}

We illustrate the three-stage text-to-IDSL parsing pipeline (Section~\ref{subsubsec:Text Description Parsing}) with a concrete example. Given the following user description:

\begin{quote}
\textit{``I want a modern three-bedroom apartment. The living room should have a sofa facing the TV stand, with a coffee table in between. The dining room is adjacent to the living room and contains a table with four chairs. The kitchen connects to the dining room through an open passage.''}
\end{quote}

\paragraph{Step 1: Building-Level Parsing.}
GPT-4 extracts the global scene structure including all room contours with consistent shared edges, and structural openings (doors, windows, open passages):

\begin{lstlisting}[language=json, style=idslstyle, caption={Building-level IDSL output.}, label={lst:building-level}]
{
  "building": {
    "building_id": "suite_001",
    "scene_tags": ["Semantics(building)"],
    "room_entities": [
      "living-room_0/0", "dining-room_0/0",
      "kitchen_0/0", "bedroom_0/0",
      "bedroom_0/1", "bathroom_0/0",
      "bathroom_0/1", "bedroom_0/2"
    ],
    "relations": [{
      "neighbours": [
        [1,8], [0,2], [1,3,4],
        [2,6], [2,5], [4,7], [3], [5], [0]
      ],
      "rooms": [
        "living-room_0/0", "dining-room_0/0",
        "kitchen_0/0", "bedroom_0/0",
        "bedroom_0/1", "bathroom_0/0",
        "bathroom_0/1", "bedroom_0/2",
        "exterior_0/0"
      ],
      "entrance": 0
    }]
  },
  "rooms": {
    "living-room_0/0": {
      "room_type": "living-room",
      "room_contour": [
        [7.5,1.0],[14.0,1.0],[14.0,7.5],
        [7.5,7.5],[7.5,1.0]
      ],
      "area": 42.25,
      "bounds": [7.5,1.0,14.0,7.5],
      "room_dimensions": {
        "width":6.5,"length":6.5,"height":3.0
      },
      "room_tags": ["Semantics(room)",
        "Semantics(living-room)"],
      "room_relations": [{
        "relation_type": "SharedEdge",
        "target_room": "dining-room_0/0"
      }]
    },
    "dining-room_0/0": {
      "room_type": "dining-room",
      "room_contour": [
        [1.5,1.5],[7.5,1.5],[7.5,7.5],
        [1.5,7.5],[1.5,1.5]
      ],
      "area": 36.0,
      "bounds": [1.5,1.5,7.5,7.5],
      "room_dimensions": {
        "width":6.0,"length":6.0,"height":3.0
      },
      "room_tags": ["Semantics(room)",
        "Semantics(dining-room)"],
      "room_relations": [{
        "relation_type": "SharedEdge",
        "target_room": "living-room_0/0"
      },{
        "relation_type": "SharedEdge",
        "target_room": "kitchen_0/0"
      }]
    }
    // ... remaining rooms omitted for brevity
  },
  "openings": {
    "door_0": {
      "semantic_tags": [
        "Semantics(cutter)", "Semantics(door)"
      ],
      "position_world": [7.5, 5.457, 1.146],
      "relation_graph": [{
        "relation": {"relation_type": "CutFrom"},
        "target_name": "dining-room_0/0"
      },{
        "relation": {"relation_type": "CutFrom"},
        "target_name": "living-room_0/0"
      }]
    },
    "open_0": {
      "semantic_tags": [
        "Semantics(cutter)", "Semantics(open)"
      ],
      "position_world": [4.0, 7.5, 1.539],
      "relation_graph": [{
        "relation": {"relation_type": "CutFrom"},
        "target_name": "dining-room_0/0"
      },{
        "relation": {"relation_type": "CutFrom"},
        "target_name": "kitchen_0/0"
      }]
    },
    "entrance_0": {
      "semantic_tags": [
        "Semantics(cutter)", "Semantics(door)"
      ],
      "position_world": [9.871, 1.0, 1.146],
      "relation_graph": [{
        "relation": {"relation_type": "CutFrom"},
        "target_name": "living-room_0/0"
      },{
        "relation": {"relation_type": "CutFrom"},
        "target_name": "exterior_0/0"
      }]
    },
    "window_0": {
      "semantic_tags": [
        "Semantics(cutter)", "Semantics(window)"
      ],
      "position_world": [4.119, 1.5, 1.803],
      "relation_graph": [{
        "relation": {"relation_type": "CutFrom"},
        "target_name": "dining-room_0/0"
      }]
    }
    // ... remaining openings omitted for brevity
  },
  "room_segments": {
    "living-room_0/0": "The living room should
      have a sofa facing the TV stand, with a
      coffee table in between.",
    "dining-room_0/0": "The dining room is adjacent
      to the living room and contains a table with
      four chairs.",
    "kitchen_0/0": "The kitchen connects to the
      dining room through an open passage."
    // ... remaining segments omitted
  }
}
\end{lstlisting}

Note that \texttt{living-room\_0/0} and \texttt{dining-room\_0/0} share the edge at $x=7.5$, ensuring geometric consistency. The \texttt{open\_0} entry represents the open passage between the dining room and kitchen. The \texttt{entrance\_0} connects \texttt{living-room\_0/0} to \texttt{exterior\_0/0}, identifying the living room as the entrance room (index 0).

\paragraph{Step 2: Room-Level Parsing.}
For each room, GPT-4 receives the fixed room geometry and generates a furniture plan. Below we show the plan for \texttt{dining-room\_0/0}:

\begin{lstlisting}[language=json, style=idslstyle, caption={Room-level furniture plan for the dining room.}, label={lst:room-plan}]
{
  "furniture_plan": [
    {
      "object_name": "DiningTable_001",
      "category": "diningtable",
      "semantic_tags": ["Semantics(furniture)",
        "Semantics(table)"],
      "placement_preference": "in the center",
      "wall_adjacent": false
    },
    {
      "object_name": "Chair_001",
      "category": "chair",
      "semantic_tags": ["Semantics(seating)",
        "Semantics(furniture)"],
      "placement_preference": "around the table",
      "wall_adjacent": false
    },
    {
      "object_name": "Chair_002",
      "category": "chair",
      "semantic_tags": ["Semantics(seating)",
        "Semantics(furniture)"],
      "placement_preference": "around the table",
      "wall_adjacent": false
    },
    {
      "object_name": "Chair_003",
      "category": "chair",
      "semantic_tags": ["Semantics(seating)",
        "Semantics(furniture)"],
      "placement_preference": "around the table",
      "wall_adjacent": false
    },
    {
      "object_name": "Chair_004",
      "category": "chair",
      "semantic_tags": ["Semantics(seating)",
        "Semantics(furniture)"],
      "placement_preference": "around the table",
      "wall_adjacent": false
    },
    {
      "object_name": "LargeShelf_001",
      "category": "largeshelf",
      "semantic_tags": ["Semantics(furniture)",
        "Semantics(storage)"],
      "placement_preference":
        "against the wall near the window",
      "wall_adjacent": true
    }
  ]
}
\end{lstlisting}

The furniture plan captures \emph{what} to place and \emph{where conceptually} (via \texttt{placement\_preference}), without committing to precise coordinates. This intermediate representation allows the object-level stage to focus purely on spatial arrangement.

\paragraph{Step 3: Object-Level Parsing.}
For each room, GPT-4 receives the fixed geometry, openings, and furniture plan, and generates precise spatial configurations. Below we show the object-level output for \texttt{dining-room\_0/0}:

\begin{lstlisting}[language=json, style=idslstyle, caption={Object-level IDSL output for the dining room, showing relation constraint instantiation.}, label={lst:obj-level}]
{
  "objs": {
    "DiningTable_001": {
      "category": "diningtable",
      "semantic_tags": ["Semantics(furniture)",
        "Semantics(table)", "Semantics(object)"],
      "position_world": [4.5, 4.5, 0.114],
      "rotation_euler": [0.0, 0.0, 0.0],
      "bounding_box": {
        "min": [3.5,3.5,0.114],
        "max": [5.5,5.5,0.874],
        "center": [4.5,4.5,0.494],
        "size": [2.0,2.0,0.760]
      },
      "relation_graph": [{
        "relation": {
          "relation_type": "StableAgainst",
          "margin": 0.0, "check_z": true,
          "child_tags": ["Subpart(bottom)"],
          "parent_tags": ["Subpart(support)"]
        },
        "target_name": "dining-room_0/0",
        "child_plane_idx": 0,
        "parent_plane_idx": 0
      }]
    },
    "Chair_001": {
      "category": "chair",
      "semantic_tags": ["Semantics(seating)",
        "Semantics(furniture)", "Semantics(chair)"],
      "position_world": [3.134, 3.641, 0.573],
      "rotation_euler": [0.0, 0.0, -2.356],
      "bounding_box": {
        "min": [2.548,3.055,0.114],
        "max": [3.427,3.934,1.028],
        "center": [2.988,3.495,0.571],
        "size": [0.879,0.879,0.914]
      },
      "relation_graph": [{
        "relation": {
          "relation_type": "StableAgainst",
          "margin": 0.0, "check_z": true,
          "child_tags": ["Subpart(bottom)"],
          "parent_tags": ["Subpart(support)"]
        },
        "target_name": "dining-room_0/0",
        "child_plane_idx": 0,
        "parent_plane_idx": 1
      }]
    },
    "Chair_002": {
      "category": "chair",
      "semantic_tags": ["Semantics(seating)",
        "Semantics(furniture)", "Semantics(chair)"],
      "position_world": [5.785, 5.763, 0.573],
      "rotation_euler": [0.0, 0.0, 1.571],
      "bounding_box": {
        "min": [5.449,5.684,0.114],
        "max": [6.121,6.254,1.028],
        "center": [5.785,5.969,0.571],
        "size": [0.672,0.570,0.914]
      },
      "relation_graph": [{
        "relation": {
          "relation_type": "StableAgainst",
          "margin": 0.0, "check_z": true,
          "child_tags": ["Subpart(bottom)"],
          "parent_tags": ["Subpart(support)"]
        },
        "target_name": "dining-room_0/0",
        "child_plane_idx": 0,
        "parent_plane_idx": 4
      }]
    },
    "Chair_003": {
      "category": "chair",
      "semantic_tags": ["Semantics(seating)",
        "Semantics(furniture)", "Semantics(chair)"],
      "position_world": [6.956, 6.489, 0.573],
      "rotation_euler": [0.0, 0.0, -1.571],
      "bounding_box": {
        "min": [6.619,5.998,0.114],
        "max": [7.292,6.568,1.026],
        "center": [6.956,6.283,0.570],
        "size": [0.673,0.570,0.912]
      },
      "relation_graph": [{
        "relation": {
          "relation_type": "StableAgainst",
          "margin": 0.0, "check_z": true,
          "child_tags": ["Subpart(bottom)"],
          "parent_tags": ["Subpart(support)"]
        },
        "target_name": "dining-room_0/0",
        "child_plane_idx": 0,
        "parent_plane_idx": 1
      }]
    },
    "Chair_004": {
      "category": "chair",
      "semantic_tags": ["Semantics(seating)",
        "Semantics(furniture)", "Semantics(chair)"],
      "position_world": [5.796, 2.029, 0.573],
      "rotation_euler": [0.0, 0.0, -3.142],
      "bounding_box": {
        "min": [5.305,1.692,0.114],
        "max": [5.875,2.365,1.029],
        "center": [5.590,2.029,0.572],
        "size": [0.570,0.673,0.915]
      },
      "relation_graph": [{
        "relation": {
          "relation_type": "StableAgainst",
          "margin": 0.0, "check_z": true,
          "child_tags": ["Subpart(bottom)"],
          "parent_tags": ["Subpart(support)"]
        },
        "target_name": "dining-room_0/0",
        "child_plane_idx": 0,
        "parent_plane_idx": 1
      }]
    },
    "LargeShelf_001": {
      "category": "largeshelf",
      "semantic_tags": ["Semantics(furniture)",
        "Semantics(storage)", "Semantics(object)"],
      "position_world": [6.652, 1.829, 0.114],
      "rotation_euler": [0.0, 0.0, 1.571],
      "bounding_box": {
        "min": [6.024,1.684,0.114],
        "max": [7.279,1.967,1.977],
        "center": [6.652,1.826,1.046],
        "size": [1.255,0.283,1.863]
      },
      "relation_graph": [{
        "relation": {
          "relation_type": "StableAgainst",
          "margin": 0.0, "check_z": true,
          "child_tags": ["Subpart(bottom)"],
          "parent_tags": ["Subpart(support)"]
        },
        "target_name": "dining-room_0/0",
        "child_plane_idx": 0,
        "parent_plane_idx": 3
      },{
        "relation": {
          "relation_type": "StableAgainst",
          "margin": 0.07, "check_z": true,
          "child_tags": ["Subpart(back)"],
          "parent_tags": ["Subpart(wall)"]
        },
        "target_name": "dining-room_0/0",
        "child_plane_idx": 0,
        "parent_plane_idx": 2
      }]
    }
  }
}
\end{lstlisting}

The example above demonstrates how relation constraints are instantiated in practice:
\begin{itemize}
    \item \textbf{Floor support}: \texttt{DiningTable\_001} and all four \texttt{Chair} instances each have a \texttt{StableAgainst} relation with \texttt{child\_ta \\ gs: ["Subpart(bottom)"]} and \texttt{parent\_tags: ["Subpart \\ (support)"]}, encoding that their bottom surfaces rest on the room's floor.
    \item \textbf{Wall leaning}: \texttt{LargeShelf\_001} has a second \texttt{StableAgain \\ st} relation with \texttt{child\_tags: ["Subpart(back)"]} and \texttt{parent\_tags: ["Subpart(wall)"]}, with a margin of 0.07, encoding that its back surface leans against the room's wall with a small gap. This directly instantiates the \texttt{placement\_ \\ preference: "against the wall near the window"} from the furniture plan.
    \item \textbf{Structural openings}: The \texttt{CutFrom} relations in Step~1 encode which rooms each door, window, or open passage connects, enabling the optimization (Section~\ref{subsec:optimization}) to enforce opening placement on the correct wall segments.
\end{itemize}

This three-stage example demonstrates how the hierarchical parsing pipeline progressively converts a free-form textual description into a complete IDSL configuration: Step~1 establishes global structure and geometry, Step~2 plans room-level furniture composition, and Step~3 determines precise spatial arrangements with explicit relation constraints. Each stage builds upon the previous, with downstream optimization further refining the initial estimates.

\subsection{CAD Floor Plan Parsing Details}
\label{app:cad-parsing}

This appendix details the geometric parsing pipeline for CAD floor plan inputs (Section~\ref{subsubsec:CAD Floor Plan Parsing}). The input is an SVG file following the format of CubiCasa5K~\cite{kalervo2019cubicasa5k}, where architectural elements are organized as semantically labeled \texttt{<g>} groups with class attributes indicating their type (\texttt{Space}, \texttt{Door}, or \texttt{Window}).

\subsubsection{Geometric Extraction}

The parser traverses the SVG DOM tree and identifies groups by their class attributes. For each \texttt{Space} group, all child \texttt{<polygon>} and \texttt{<polyline>} elements are collected. Their vertex coordinates are transformed from local to world coordinates by composing the SVG transform matrices (supporting \texttt{matrix}, \texttt{translate}, \texttt{rotate}, and \texttt{scale} operations) along the ancestor chain from the root to the current element. When a \texttt{Space} group contains multiple polygons, the one with the largest area is selected as the room contour, as smaller polygons typically correspond to annotation artifacts. Room types (e.g., \texttt{LivingRoom}, \texttt{Bedroom}, \texttt{Bath}) are extracted from the class attributes of each \texttt{Space} group and mapped to canonical IDSL labels (e.g., \texttt{LivingRoom} $\rightarrow$ \texttt{living-room\_0/0}).

For \texttt{Door} and \texttt{Window} groups, the same coordinate transformation is applied to all child polygon vertices. The centroid of all transformed vertices within each group is computed as the representative position of that door or window.

\subsubsection{Geometric Refinement}

\paragraph{Vertex Snapping.}
Raw room contours extracted from SVG often exhibit small coordinate misalignments at shared boundaries due to drawing imprecision. To resolve this, we perform global vertex snapping: all room vertices are collected, and pairs within a distance threshold are merged into clusters using a union-find structure. Each cluster is replaced by its centroid coordinate, ensuring that adjacent rooms share geometrically identical wall vertices. Degenerate polygons (area $<$ threshold) produced by snapping are discarded and replaced with the original polygon.

\paragraph{Shared Edge Computation.}
After snapping, shared edges between adjacent rooms are computed using Shapely's polygon boundary intersection. For each pair of rooms, the intersection of their boundary geometries yields the set of shared line segments. These segments are merged via \texttt{linemerge} to produce continuous shared edges. A post-processing step straightens near-axis-aligned triangle artifacts, which arise when adjacent room contours have slightly misaligned intermediate vertices along a shared wall: for 3-vertex polylines whose endpoints are approximately axis-aligned, the middle vertex is projected onto the line connecting the endpoints.

\paragraph{Door/Window Matching.}
Each door and window centroid is matched to its adjacent rooms by computing the point-to-segment distance from the centroid to every edge of every room contour. The two rooms with the smallest boundary distances are assigned as the adjacent rooms for that opening.

\subsubsection{IDSL Assembly}

The extracted and refined geometric data is assembled into a complete IDSL configuration:

\begin{itemize}
    \item Building level: The room entity list and inter-room adjacency graph are constructed from the shared-edge computation. The \texttt{neighbours} list records which rooms share edges. The entrance room is identified as the room whose door has a \texttt{CutFrom} relation to \texttt{exterior\_0/0}.
    \item Room level: Each room is stored with its snapped contour coordinates, bounding box, area, room type, and \texttt{SharedEdge} relations to adjacent rooms.
    \item Structural elements: Doors and windows are stored with their world-space positions and \texttt{CutFrom} relations linking them to their two adjacent rooms.
    \item Object level: Since CAD floor plans typically do not contain detailed furniture information, object-level content is generated by invoking the room-level LLM prompt (Section~\ref{subsubsec:Text Description Parsing}), with the room description replaced by the room type as the generation cue. The LLM infers typical furniture configurations based on room function and dimensions.
\end{itemize}

The resulting IDSL configuration $\mathcal{S}_0$ preserves the geometric precision of the original floor plan while augmenting it with semantic annotations and furniture layouts, providing a complete scene specification for optimization (Section~\ref{subsec:optimization}).

\subsection{CAD-to-scene Capability}
\label{subsec:cad_verify}

\begin{table}[t]
\centering
\caption{CAD-to-scene Capability. Higher is better for both metrics.}
\label{tab:cad_verify_results}
\resizebox{0.75\linewidth}{!}{
\begin{tabular}{lcc}
\toprule
\textbf{Method} & \textbf{Room IoU} $\uparrow$ & \textbf{Adjacency F1} $\uparrow$ \\
\midrule
RoomPilot & 0.83 & 0.85\\
\bottomrule
\end{tabular}}
\end{table}

In this section, we evaluate the CAD-to-scene capability of RoomPilot under multi-room CAD floor-plan inputs.
This experiment isolates the CAD parsing pathway and quantifies how well the generated scene structure matches the input floor plan in terms of both geometry and room connectivity.

Given a multi-room CAD floor plan as input, RoomPilot reconstructs a multi-room indoor scene represented as a set of room regions and their pairwise adjacency relations.
For quantitative evaluation, we derive a reference scene structure from the CAD input, where each room is represented by a 2D polygon and inter-room connectivity is defined by shared architectural boundaries.
We compare the generated scene structure against this CAD-derived reference in terms of room geometry and room adjacency.

We report two structure-oriented metrics defined as follows:
\begin{itemize}
    \item Room IoU.
    Let $\mathcal{R} = \{R_i\}_{i=1}^{N}$ denote the set of reference room regions derived from the CAD floor plan, and $\hat{\mathcal{R}} = \{\hat{R}_i\}_{i=1}^{N}$ the corresponding room regions produced by the generated scene structure.
    After projecting both structures onto the floor plane, the Room IoU is defined as
    \begin{equation}
        \mathrm{Room\ IoU}
        =
        \frac{1}{N}
        \sum_{i=1}^{N}
        \frac{|R_i \cap \hat{R}_i|}{|R_i \cup \hat{R}_i|},
    \end{equation}
    which measures the geometric alignment between generated room regions and the CAD-derived reference.

    \item Adjacency F1.
    We construct an undirected room adjacency graph $G = (\mathcal{V}, \mathcal{E})$ from the reference structure, where each node corresponds to a room and an edge $(i,j) \in \mathcal{E}$ indicates that rooms $i$ and $j$ are directly connected in the CAD floor plan.
    Similarly, we extract a predicted adjacency graph $\hat{G} = (\mathcal{V}, \hat{\mathcal{E}})$ from the generated scene structure.
    The Adjacency F1 score is computed based on edge-level precision and recall:
    \begin{equation}
        \mathrm{Precision} = \frac{|\mathcal{E} \cap \hat{\mathcal{E}}|}{|\hat{\mathcal{E}}|},
        \quad
        \mathrm{Recall} = \frac{|\mathcal{E} \cap \hat{\mathcal{E}}|}{|\mathcal{E}|},
    \end{equation}
    \begin{equation}
        \mathrm{Adjacency\ F1}
        =
        \frac{2 \cdot \mathrm{Precision} \cdot \mathrm{Recall}}
        {\mathrm{Precision} + \mathrm{Recall}}.
    \end{equation}
    This metric evaluates the consistency of inter-room connectivity between the generated structure and the CAD-derived reference.
\end{itemize}

Table~\ref{tab:cad_verify_results} reports the quantitative results.
RoomPilot achieves high Room IoU and Adjacency F1 under multi-room CAD conditioning, indicating that the generated structure remains well aligned with the input floor plan in both geometry and connectivity.

\subsection{Cross-Modal Semantic Parsing to IDSL}

As illustrated in Fig.~\ref{fig:overview}, we unify heterogeneous inputs into a common scene representation through IDSL, which serves as the sole interface between parsing and optimization. Given either a free-form textual description or a CAD floor plan, the system produces an initial IDSL state \( \mathcal{S}_0 \) that explicitly encodes rooms, objects, semantic attributes, and spatial or functional relations. Neither the language model nor the CAD parser interacts directly with the solver; all downstream optimization operates exclusively on IDSL.

For textual input, the LLM is used only at the parsing stage to transform free-form language into structured IDSL specifications. Specifically, the LLM is guided by a set of fixed \emph{IDSL prompts} (shown as \textit{Prompts for IDSL} in Fig.~\ref{fig:overview}), which define a constrained output schema aligned with IDSL primitives. These prompts instruct the model to extract entities (rooms, objects, attributes) and relations (spatial, functional, hierarchical) and to emit them in a normalized, machine-readable form rather than natural language. The resulting structured output is then deterministically converted into IDSL constructs with symbolic rules and parametric preferences, forming \( \mathcal{S}_0 \). This prompting process is a one-shot semantic grounding step and does not involve any iterative reasoning or optimization.

For CAD floor plans, room contours, walls, doors, and windows are parsed using rule-based geometric procedures and directly mapped into IDSL with explicit polygonal geometry and topological relations. Regardless of whether \( \mathcal{S}_0 \) originates from LLM-based parsing or CAD-based parsing, the solver refines the scene solely through structured energy minimization defined over IDSL, making the overall pipeline modality-agnostic, deterministic, and fully decoupled.

\section{Details of Self-Regulating Scene Optimization}
\label{sec:opt_supp}

This section provides additional detail on the optimization process outlined in
Section~3.3. While Appendix \ref{sec:idsl_supp} specifies the IDSL representation, the optimizer governs
how an initial configuration evolves into a structurally valid and semantically
organized scene. Algorithm~\ref{alg:self_regulating} summarizes the complete control
flow; the paragraphs below clarify how each component interacts with the IDSL state.

\paragraph{Optimization Variables.}
The optimizer acts directly on explicit IDSL fields. At the Building Level, the global
footprint and room-connectivity graph constrain admissible configurations. Room-Level
attributes—room contours, shared edges, and derived spatial measures—regulate object
placement and prevent geometric drift. The most active variables lie at the Object
Level: world-space pose (position, rotation, scale), affine basis, rotation axis,
semantic tags, relation-graph entries, and global/local bounding boxes. These fields
form the primary degrees of freedom for structural repair and semantic refinement.

\paragraph{Dual-Channel Energy.}
Structural and semantic energies, defined in Eq.~(1), evaluate complementary aspects of
the IDSL state. Structural terms assess geometric validity (collisions, support,
alignment, circulation). Semantic terms measure functional relations, grouping
patterns, and room-consistency. Because IDSL stores all relevant attributes explicitly,
both energies can be computed incrementally and locally, yielding a transparent and
well-conditioned energy landscape.

\paragraph{Proposal Operators Across Levels.}
Candidate updates $\mathcal{S}'$ are generated by operators acting at different IDSL
levels. Building-Level operators ensure consistency of global topology. Room-Level
operators refine boundary relations when needed. Object-Level operators dominate the
search: translations within room polygons, axis-aligned or affine-basis rotations,
scale adjustments, and relation-preserving moves that maintain stability or in-room
membership. Explicit relational fields allow invalid proposals to be rejected early,
reducing unnecessary energy evaluations.

\paragraph{Staged Rule Activation.}
The grouped rule schedule ($R^{(1)},\dots,R^{(L)}$) aligns with the three IDSL levels.
Early stages enforce structural validity; middle stages activate room functions and
object-group semantics; later stages refine orientation cues and stylistic relations.
This progression prevents early coupling of heterogeneous constraints and produces a
stable structural–functional–semantic optimization path.

\paragraph{Self-Regulating Annealing.}
Algorithm~\ref{alg:self_regulating} uses the rule-alignment score $\rho(t)$ as a single
feedback signal that adjusts both the temperature (Eq.~5) and operator probabilities
(Eq.~4). When violations are common, the system favors exploratory moves; as the scene
becomes consistent with the active rule set, the temperature decreases and the operator
distribution shifts toward conservative refinements. Because $\rho(t)$ is computed
directly from IDSL fields, the annealing behavior faithfully reflects scene structure.

\paragraph{Local Evaluation and Acceptance.}
Energy differences are evaluated only over modified objects and their geometric or
relational neighbors, reducing computational cost while preserving accuracy.
Structural improvements ($\Delta E_{\mathrm{struct}}<0$) are always accepted, ensuring
monotonic geometric convergence. Semantic improvements are accepted using a
temperature-controlled rule, allowing flexibility early in the search and determinism
as the system stabilizes.

\paragraph{Convergence.}
The combination of explicit IDSL attributes, staged rule activation, adaptive
annealing, and localized evaluation yields consistent convergence even in multi-room
scenes. Global constraints from the Building and Room Levels prevent drift, while
Object-Level fields support fine-grained semantic adjustments without compromising
structural feasibility. The system therefore follows a coherent evolution from coarse
structural repair to functional organization and finally semantic refinement.

\begin{algorithm}[t]
\caption{Self-Regulating Scene Optimization}
\label{alg:self_regulating}
\begin{algorithmic}

\Require Initial IDSL state $\mathcal{S}_0$; rule groups $\{R^{(l)}\}_{l=1}^L$;
operator set $\{\Omega_i\}$ with initial weights $\{\beta_i^{(0)}\}$;
initial temperature $T_0$

\State $\mathcal{S} \gets \mathcal{S}_0$
\State $\kappa \gets 1$

% \For{$t = 1$ to $T_{\max}$}
\For{$t \gets 1$ \textbf{to} $T_{\max}$}

    \State $\rho(t) \gets \text{ComputeRuleAlignment}(\mathcal{S}, R^{(\kappa)})$

    \If{$\mathcal{S}$ is stable under $R^{(\kappa)}$}
        \State $\kappa \gets \min(\kappa + 1, L)$
    \EndIf

    \State $R(t) \gets \text{ActivateRuleSet}(\kappa)$
    \State Update $\alpha_{\mathrm{struct}}(t)$ and $\alpha_{\mathrm{sem}}(t)$
    \State Update $\beta_i(t)$ and $p_i(t)$ using Eq.~(4)
    \State Update $T(t)$ using Eq.~(5)

    \State Sample operator index $i \sim p_i(t)$
    \State $\mathcal{S}' \gets \Omega_i(\mathcal{S})$

    \State $N \gets \text{ModifiedSet}(\mathcal{S}, \mathcal{S}')$
    \State $\text{adj}(N) \gets \text{Neighborhood}(N)$

    \State Compute $\Delta E_{\mathrm{struct}}$ and $\Delta E_{\mathrm{sem}}$ on $N \cup \text{adj}(N)$

    \If{$\Delta E_{\mathrm{struct}} < 0$}
        \State $\mathcal{S} \gets \mathcal{S}'$
    \ElsIf{structural feasibility is preserved}
        \State Accept $\mathcal{S}'$ with probability $P_{\mathrm{acc}}(\Delta E_{\mathrm{sem}}, T(t))$
    \EndIf

\EndFor

\State \Return $\mathcal{S}$

\end{algorithmic}
\end{algorithm}

\begin{table*}[t]
\centering
\caption{Fields and optimization roles at the \textbf{Building Level} of IDSL,
describing topology, connectivity, and footprint.}
\resizebox{\textwidth}{!}{
\begin{tabular}{lccc}
\toprule
Field & Type & Description & Role in Optimization \\
\midrule
\texttt{building\_id} & string &
Unique identifier of the building or suite. &
Used for global indexing and energy aggregation. \\
\texttt{floor\_outline} & array[float] &
2D polygon describing the global building footprint. &
Defines the domain boundary for layout search space. \\
\texttt{room\_entities} & list &
List of all rooms contained in this building. &
Defines the nodes of the room-level connectivity graph. \\
\texttt{scene\_tags} & list[string] &
High-level semantic tags (e.g., ``Semantics(building)''). &
Activates global structural and semantic energy terms. \\
\texttt{relations} & list[dict] &
Building-level relations (e.g., between circulation spaces). &
Provides optional constraints for global connectivity. \\
\bottomrule
\end{tabular}}
\label{tab:idsl_building}
\end{table*}

\begin{table*}[t]
\centering
\caption{Fields and optimization roles at the \textbf{Room Level} of IDSL,
describing geometry, semantics, and boundaries.}
\resizebox{\textwidth}{!}{
\begin{tabular}{lccc}
\toprule
Field & Type & Description & Role in Optimization \\
\midrule
\texttt{room\_id} & string &
Unique identifier of the room instance. &
Node identifier in the structural--semantic graph. \\
\texttt{room\_type} & string &
Functional type (e.g., \texttt{living\_room}, \texttt{bedroom}). &
Determines semantic priors for object arrangement. \\
\texttt{room\_contour} & array[float] &
2D polygon defining the room footprint. &
Provides the geometric boundary for object placement. \\
\texttt{room\_tags} & list[string] &
Semantic attributes of the room (e.g., style, function). &
Guides semantic compatibility and zoning constraints. \\
\texttt{room\_relations} & list[dict] &
Adjacency and shared-boundary relations to other rooms &
Enforces coherent inter-room alignment and connectivity. \\
\texttt{active} & bool &
Indicates whether this room participates in the current iteration. &
Supports staged and progressive optimization scheduling. \\
\bottomrule
\end{tabular}}
\label{tab:idsl_room}
\end{table*}

\begin{table*}[t]
\centering
\caption{Fields and optimization roles at the \textbf{Object Level} of IDSL, aligned with the actual JSON schema, including affine and rotational DOF fields.}
\resizebox{\textwidth}{!}{
\begin{tabular}{lccc}
\toprule
Field & Type & Description & Role in Optimization \\
\midrule
\texttt{object\_id} & string &
Unique identifier of the object instance. &
Index key in the object-level optimization graph. \\

\texttt{polygon} & array / null &
Optional 2D footprint for planar alignment. &
Used when evaluating wall or floor contact. \\

\texttt{semantic\_tags} & list[string] &
Semantic annotations (e.g., \texttt{Semantics(furniture)}). &
Activates semantic rules and compatibility constraints. \\

\texttt{relation\_graph} & list[dict] &
Relations to rooms and other objects (e.g., \texttt{StableAgainst}). &
Core input to relational energy terms in $E_{\mathrm{sem}}$. \\

\texttt{position\_world} & array[3] &
Global 3D coordinates of the object center. &
Primary variable for spatial placement optimization. \\

\texttt{rotation\_euler} / \texttt{rotation\_quaternion} & array &
Orientation of the object in world space. &
Pose variables for orientation alignment. \\

\texttt{rotation\_axis} & vector3 &
Canonical rotation axis (from DOF specification). &
Constrains allowable rotational degrees of freedom. \\

\texttt{scale} & array[3] &
Scaling factors along x/y/z axes. &
Allows mild resizing to fit the spatial envelope. \\

\texttt{affine\_basis} & matrix(3$\times$3) &
Local affine translation basis (from DOF matrix). &
Defines local coordinate frame for geometric optimization. \\

\texttt{transform\_matrix} & matrix(4$\times$4) &
Transformation from local to world coordinates. &
Used to compute global geometry during optimization. \\

\texttt{bounding\_box} & dict &
\{\texttt{min}, \texttt{max}, \texttt{center}, \texttt{size}\}. &
Used for collision detection and volumetric energy terms. \\

\texttt{bbox\_corners} & array[8][3] &
World-space coordinates of bounding-box corners. &
Enables precise contact and overlap evaluation. \\

\texttt{local\_bbox} & dict &
Local-coordinate bounding box (center and size). &
Maintains intra-object geometric consistency. \\

\texttt{active} & bool &
Object participation flag. &
Controls staged optimization and pruning. \\
\bottomrule
\end{tabular}}
\label{tab:idsl_object}
\end{table*}

\section{Details of IDSL Specification}
\label{sec:idsl_supp}

The IDSL provides the unified, hierarchical scene
representation used throughout multimodal parsing and self-regulating optimization framework. While Section~3.2 introduces IDSL conceptually, this section expands upon its formal structure. IDSL organizes all geometric, semantic, and relational attributes across three complementary levels---\textbf{Building}, \textbf{Room},
and \textbf{Object}---each encoding a distinct layer of spatial abstraction. Tables~\ref{tab:idsl_building},
\ref{tab:idsl_room}, and \ref{tab:idsl_object} define the complete field specifications
for these three levels.

\subsection{Building Level}

The Building Level defines the \emph{global structural backbone} of the environment.
It captures the geometric extent, the set of room entities, and the topological
relationship among them.

\paragraph{Global footprint.}
The building boundary is represented as a 2D polygon (\texttt{floor\_outline}), forming
the domain within which all layout and optimization operations occur. This polygon
supports arbitrarily complex outlines, including concave shapes.

\paragraph{Room composition.}
The building maintains an explicit list of all room identifiers (\texttt{room\_entities}),
which enables stable indexing, room-level iteration, and consistent association
between global topology and local room attributes.

\paragraph{Connectivity graph.}
Instead of pairwise room relations, IDSL uses a compact adjacency graph to encode
inter-room connectivity:
\begin{itemize}
  \item nodes correspond to room identifiers,
  \item edges represent physical adjacency or circulation paths,
  \item an optional entrance index specifies the primary access point.
\end{itemize}
This minimal and unambiguous structure provides the necessary constraints for preserving
architectural feasibility and maintaining circulation consistency during optimization.

\paragraph{Semantic annotations.}
High-level building tags (e.g., \texttt{Semantics( \\ building)}) allow the optimizer to
distinguish global structural constraints from room- and object-level constraints.

\subsection{Room Level}

The Room Level encodes the \emph{spatial identity and functional semantics} of each
individual room. It forms the intermediate layer that links the building-level topology
with object-level placement.

\paragraph{Polygonal room contour.}
Each room is specified by a 2D polygon (\texttt{polygon\_coords}) describing its
footprint. This representation accommodates convex, concave, or irregular room shapes
and allows precise boundary-aware placement of furniture.

\paragraph{Room semantics.}
Semantic tags distinguish the functional role of each room (e.g.,
\texttt{Semantics(Bedroom)}, \texttt{Semantics(Kitchen)}), and support mappings from
natural-language descriptions to structured constraints.

\paragraph{Derived geometric properties.}
Quantities such as room area and bounding rectangle are derived from the room polygon
and used internally for structural energy terms; they are therefore not stored as
separate fields.

\paragraph{Adjacency relations.}
Shared-edge relations between rooms are explicitly recorded and provide:
\begin{itemize}
  \item boundary constraints enforcing consistent inter-room alignment,
  \item cues for door/window placement,
  \item constraints on room-wise geometric adjustments during optimization.
\end{itemize}

\subsection{Object Level}

The Object Level provides the \emph{finest-grained geometric and semantic representation}
within IDSL. Each object---furniture, fixture, or decoration---is encoded in a unified
schema, enabling consistent relational reasoning and optimization.

\paragraph{Semantic attributes.}
Objects include both high-level semantic tags (\texttt{semantic\_tags}) and a compact
object category field (\texttt{category}), which together determine the applicable
physical and semantic constraints.

\paragraph{Geometric and pose attributes.}
IDSL stores complete 3D pose information:
\begin{itemize}
  \item world-space position (\texttt{position\_world}),
  \item orientation in both Euler and quaternion form,
  \item axis-aligned scale,
  \item the full \(4\times4\) world-space transform matrix.
\end{itemize}

To support optimization, two additional fields encode local geometric degrees of freedom:
\begin{itemize}
  \item \texttt{affine\_basis}, a linear basis for local translations,
  \item \texttt{rotation\_axis}, a unit vector specifying the admissible axis of rotation.
\end{itemize}

\paragraph{Bounding-box descriptors.}
IDSL consolidates all bounding-box information into structured fields, including:
\begin{itemize}
  \item the global \texttt{bounding\_box} (with min/max/center/size),
  \item the eight world-space \texttt{bbox\_corners},
  \item the \texttt{local\_bbox} in object space.
\end{itemize}

These descriptors enable accurate evaluation of contacts, overlaps, and volumetric
constraints.

\paragraph{Relational graph.}
The \texttt{relation\_graph} records structured relationships between an object and
its surrounding entities (e.g., \emph{in-room}, \emph{on-top-of}, \emph{stable-against}).
Each relation includes a well-defined type, geometric parameters, and relevant plane
indices, supporting interpretable and optimization-ready reasoning about spatial
interaction.

\paragraph{Active-state flag.}
The \texttt{active} field allows the optimizer to dynamically include or exclude
objects during staged optimization, pruning, or scene refinement.

Together, the Building, Room, and Object levels form a unified, interpretable representation that decouples global structure from local interaction. 
An example of IDSL is provided in Listing~\ref{lst:idsl_full}
This layered organization mirrors real architectural design workflows: building-level topology, room-level organization, and object-level configuration. All optimization energies operate over these explicitly defined fields, ensuring that geometry, semantics,
and relations remain jointly consistent throughout the scene refinement process.

\begin{lstlisting}[language=json, style=idslstyle, 
caption={An IDSL example.}, 
label={lst:idsl_full}]
{
  "building": {
    "building_id": "suite_001",
    "floor_outline": [
      [1.0, 1.0], [16.0, 1.0], [16.0, 19.0], [1.0, 19.0], [1.0, 1.0]
    ],
    "scene_tags": ["Semantics(building)"],
    "room_entities": [
      "living-room_0/0", "dining-room_0/0", "kitchen_0/0",
      "bedroom_0/0", "bedroom_0/1", "bathroom_0/0",
      "bathroom_0/1", "bedroom_0/2"
    ],
    "relations": [
      {
        "neighbours": [[1,8],[0,2],[1,3,4],[2,6],[2,5],[4,7],[3],[5],[0]],
        "rooms": [
          "living-room_0/0", "dining-room_0/0", "kitchen_0/0",
          "bedroom_0/0", "bedroom_0/1", "bathroom_0/0",
          "bathroom_0/1", "bedroom_0/2", "exterior_0/0"
        ],
        "entrance": 0
      }
    ]
  },

  "rooms": {
    "dining-room_0/0": {
      "polygon_coords": [
        [1.5,1.5], [7.5,1.5], [7.5,7.5], [1.5,7.5], [1.5,1.5]
      ],
      "tags": ["Semantics(RoomContour)", "Semantics(DiningRoom)"],
      "area": 36.0,
      "bounds": [1.5,1.5,7.5,7.5],
      "relations": [
        {
          "relation_type": "SharedEdge",
          "target": "living-room_0/0",
          "value": "MULTILINESTRING ((7.5 1.5, 7.5 7.5))"
        }
      ]
    }
  },

  "objs": {
    "Chair_139218": {
      "polygon": null,
      "affine_basis": [
        [0.99999999957, -5.91e-10, -2.06e-05],
        [-5.91e-10, 0.99999999918, -2.86e-05],
        [-2.06e-05, -2.86e-05, 1.24e-09]
      ],
      "rotation_axis": [2.06e-05, 2.86e-05, 0.99999999937],
      "rotation_euler": [-2.3e-05, -1.4e-05, -2.35619],
      "rotation_quaternion": [0.38268, -1.09e-05, 7.9e-06, -0.92388],
      "scale": [1.0, 1.0, 1.0],
      "position_world": [3.1339, 3.6407, 0.5731],
      "transform_matrix": [
        [-0.7071, 0.7071, 2.6e-05, 3.1339],
        [-0.7071, -0.7071, -6.2e-06, 3.6407],
        [1.4e-05, -2.3e-05, 1.0, 0.5731],
        [0.0, 0.0, 0.0, 1.0]
      ],
      "semantic_tags": [
        "Semantics(seating)",
        "Semantics(furniture)",
        "Semantics(object)",
        "Semantics(chair)"
      ],
      "relation_graph": [
        {
          "relation": {
            "relation_type": "StableAgainst",
            "margin": 0.0,
            "check_z": true,
            "rev_normal": false
          },
          "target_name": "dining-room_0/0",
          "child_plane_idx": 0,
          "parent_plane_idx": 1,
          "value": null
        }
      ],
      "bounding_box": {
        "min": [2.5484,3.0552,0.1143],
        "max": [3.4271,3.9339,1.0281],
        "center": [2.9878,3.4946,0.5712],
        "size": [0.8787,0.8787,0.9138]
      },
      "bbox_corners": [...],
      "local_bbox": {
        "center": [0.2066, 0.0, -0.00188],
        "size": [0.5701, 0.6725, 0.9138]
      },
      "active": true,
      "category": "chair"
    }
  }
}
\end{lstlisting}

\subsection{Comparison with Infinigen Indoors.}
The Indoor Domain-Specific Language (IDSL) is designed as an intermediate, editable semantic representation that bridges user intent and scene generation.
Rather than serving as a standalone scene description format or a solver-internal execution state, IDSL functions as a human- and LLM-facing abstraction layer. It can be repeatedly modified and optimized throughout the generation pipeline. As illustrated in Figure \ref{fig:overview}, IDSL remains accessible and editable both before and after optimization, which enables iterative refinement guided by user intent.

\subsubsection{Semantic Foundations and Extensions Beyond Infinigen Indoors} 

IDSL draws inspiration from the semantic abstractions introduced in Infinigen Indoors, which provide a well-established vocabulary for describing indoor objects and their structural relationships. A subset of semantic tags and relation primitives is adopted as a starting point for representing indoor scenes.

Building upon this foundation, IDSL reorganizes and extends these semantic elements within a different representational framework. Unlike Infinigen Indoors, where such semantics are primarily used to encode final execution states, IDSL promotes them to a persistent, optimization-facing representation that remains available throughout the pipeline. This shift enables semantic information to participate in reasoning and refinement, rather than being consumed once during procedural generation.

Furthermore, IDSL enriches these semantics through user-specified constraints and is used in the optimization process. This design enables semantic information to guide spatial reasoning and iterative adjustments, achieving greater control over scene generation rather than simply serving as static annotations.

\subsubsection{Functional Role: Intermediate Language vs. Solver-Internal State}

The fundamental distinction between IDSL and Infinigen Indoors lies in how and when the representation is used. Json files from Infinigen Indoors \cite{infinigen2024indoors} represents a solver-final execution state produced after procedural generation. It encodes finalized geometric and relational information required to instantiate a scene and is not intended for iterative modification or user interaction.

By contrast, IDSL is explicitly designed to be editable, revisitable, and optimization-facing. It organizes scene information hierarchically across building, room, and object levels, enabling optimization to proceed in a staged manner while maintaining global spatial consistency. Constraints and relations can be selectively activated, relaxed, or reweighted at different stages, allowing IDSL to encode not only what the scene is, but also how it should be refined during optimization.

\subsubsection{Why IDSL Constitutes a Domain-Specific Language}

Although IDSL adopts a graph-like structure, its defining property as a domain-specific language lies in its ability to encode and manipulate user intent as structured constraints across multiple levels. In particular, IDSL supports:

\begin{itemize}
    \item Unified representation of heterogeneous inputs, enabling textual descriptions and CAD-derived layouts to be fully encoded within a single representation;
    \item Hierarchical organization across multiple levels (building, room, and object), allowing optimization to proceed in a layered manner that preserves spatial consistency;
    \item Explicit encoding of user-specified constraints, which directly influence spatial relationships during optimization rather than merely describing them.
    \item By fully encoding heterogeneous user inputs together with explicit constraints, IDSL enables a high degree of user controllability over scene generation. This design allows user intent to be consistently reflected throughout the optimization and generation process, supporting controllable and interpretable scene synthesis beyond what is achievable with purely descriptive scene graphs.
\end{itemize}

\subsubsection{Controllability Evaluation}
\label{subsec:constraint_controllability}

We compare between IDSL and Infinigen Indoors \cite{infinigen2024indoors} to evaluate controllability under explicitly specified constraints.We consider a set of indoor scenes with comparable scale and complexity. Each scene contains 5-6 rooms and approximately 10–20 objects. For IDSL, user-specified spatial and functional constraints are explicitly encoded in the intermediate representation and enforced during optimization. For Infinigen Indoors, scenes are generated using its official procedural pipeline, without introducing additional input interfaces, constraint specifications, or modifications.

For evaluation, we define a fixed set of constraints per scene, including spatial and structural predicates. Each method generates 10 scenes using different random seeds. We report the CSR\textsubscript{count} and CSR\textsubscript{rel} across generated scenes, which reflects how consistently each representation preserves user-specified constraints under stochastic generation. The CSR\textsubscript{count} and CSR\textsubscript{rel} are explained in Section \ref{sec:details of experiments_supp}.

\begin{table}[t]
\centering
\caption{Comparison of controllability performance with Infinigen Indoors \cite{infinigen2024indoors}.
% Three metrics are reported: 
% Layout Fidelity (LF, IoU-based spatial alignment), 
% Constraint Satisfaction Rate (CSR\textsubscript{count}, object quantity
}
\resizebox{0.75\linewidth}{!}{
\begin{tabular}{lcc}
\toprule
Method & 
\textbf{CSR\textsubscript{count}} $\uparrow$ & 
\textbf{CSR\textsubscript{rel}} $\uparrow$ \\
\midrule
Infinigen Indoors~\cite{infinigen2024indoors}  & 0.31 & 0.36 \\
\midrule
RoomPilot  & \textbf{0.87} & \textbf{0.80} \\
\bottomrule
\end{tabular}
}
\label{tab:constraint_csr}
\end{table}

As shown in Table~\ref{tab:constraint_csr}, RoomPilot consistently achieves higher constraint satisfaction. This indicates that explicitly encoding user intent and constraints within an intermediate, optimization-facing representation enables more predictable and controllable scene generation compared to purely procedural representations.

% \paragraph{Comparison with Infinigen Indoors.}
% Unlike the \textit{Constraint Specification API} of Infinigen Indoors, which encodes geometric and relational rules implicitly within procedural code, 
% IDSL externalizes all attributes as explicit fields in a hierarchical schema. 
% This separation of geometry, semantics, and relations across three levels allows independent yet coupled optimization through the dual-channel energy model.
% Structural terms operate on building and room geometry, while semantic and relational terms act on object-level interactions. 
% Consequently, IDSL functions as both a human-readable specification and an executable configuration for multi-room scene synthesis.

% ----------------------------------------------------------
% C.1 Apartment-Level Envelope Generation (Walls/Floors/Ceilings)
% ----------------------------------------------------------

\section{Details of Hierarchical Indoor Scene Generation}
\label{sec:generation_supp}

\subsection{Multi-level Indoor Scene Generation}

\paragraph{Apartment-Level Envelope Generation.} At the apartment level, we generate the complete architectural envelope---including walls, floors, and ceilings---directly from the spatial layouts specified in IDSL. For each room, the layout polygon is first converted into a closed baseline curve. This preprocessing enforces counter-clockwise vertex ordering, merges redundant or collinear edges, and resolves narrow-angle inconsistencies or potential T-junctions, ensuring that the curve provides a topologically stable scaffold for downstream procedural generation.

Based on this curve, we employ three procedural modules,
\[
\begin{aligned}
\mathcal{W} &= \text{PCG}_{\text{wall}}(\text{Curve}, \theta_{\text{wall}}), \\
\mathcal{F} &= \text{PCG}_{\text{floor}}(\text{Curve}, \theta_{\text{floor}}), \\
\mathcal{C} &= \text{PCG}_{\text{ceil}}(\text{Curve}, \theta_{\text{ceil}}).
\end{aligned}
\]

to generate walls, floors, and ceilings, respectively. The wall generator constructs a watertight extruded mesh controlled by structural parameters such as height, thickness, extrusion direction, and corner-joining rules, while robustly handling concave geometries and multi-room junctions. The floor and ceiling generators produce planar meshes from the same baseline curve, applying optional solidification and adaptive subdivision to ensure geometric consistency with the wall envelope.

\begin{lstlisting}[
    language=json,
    basicstyle=\small\ttfamily,
    breaklines=true,
    columns=fullflexible,
    frame=single,
    style=idslstyle,
]
{
  "material": {"category": "wall2", "material": "PRESET1"},
  "wall2": {
    "draw_direction": "CCW",
    "finish2": [
      {
        "finish_name": "brick",
        "tile_width": 0.20,
        "tile_length": 0.05,
        "spacing": 0.01,
        "mortar_depth": 0.01,
        "solidify": true,
        "thickness": 0.03,
        "pattern": "regular_tile"
      }
    ],
    ...
  }
}
\end{lstlisting}

\begin{figure}[t]
\centering
\includegraphics[width=\linewidth]{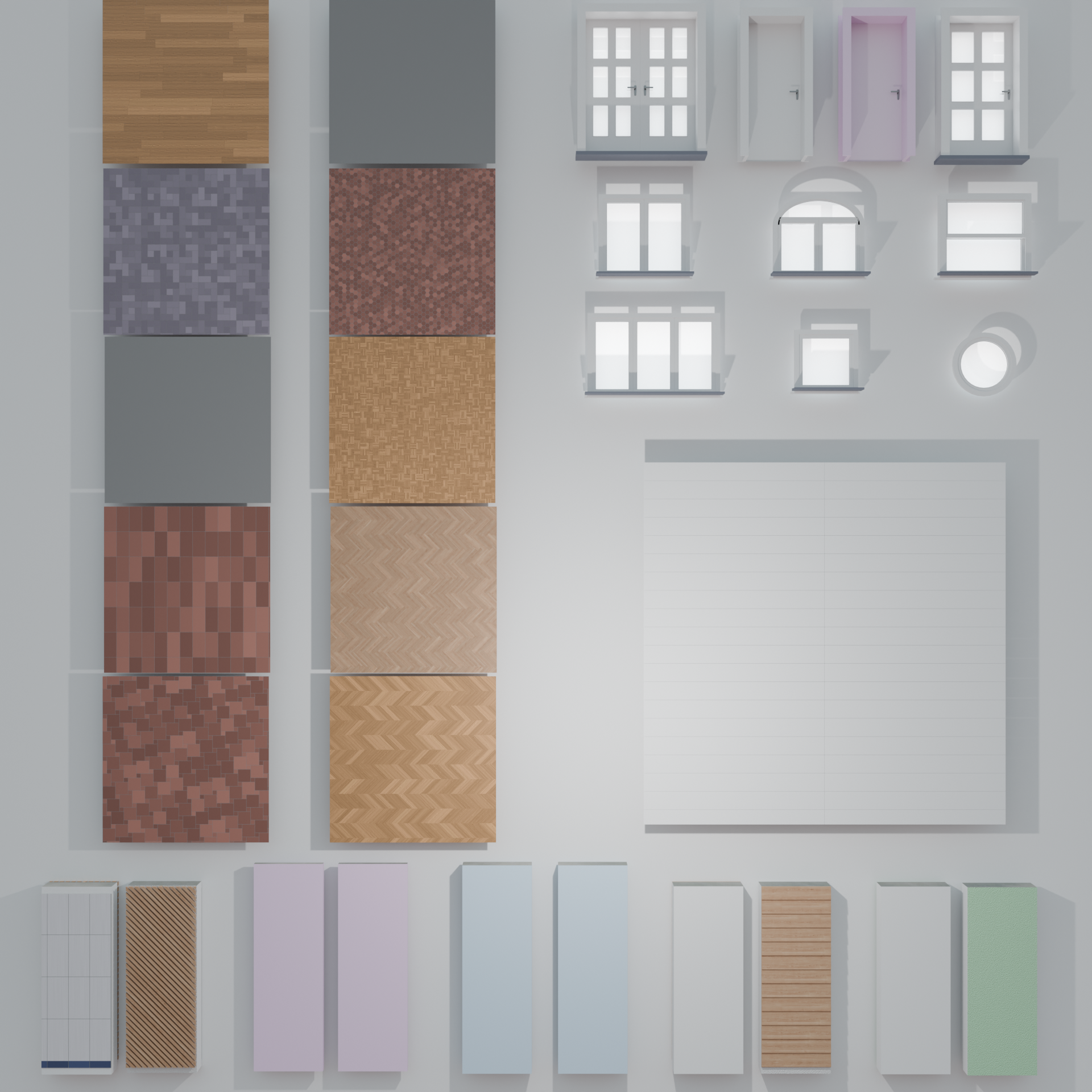}
\caption{
Visualization of the procedural presets used for structural element generation. 
Top-left: ten floor presets covering wood, tile, marble, carpet, and composite materials. 
Top-right: four door presets and six window presets illustrating diverse opening geometries and framing styles. 
Center-right: the ceiling preset used across all scenes. 
Bottom: five wall presets, where each pair shows the exterior-wall configuration on the left and the interior-wall configuration on the right.
}
\label{fig:preset_visualization}
\vspace{+0.3em}
\end{figure}

As shown in Figure \ref{fig:preset_visualization}, to support high-quality and stylistically coherent appearance, we maintain a unified library of parametric presets: \textbf{5 wall presets}, \textbf{10 floor presets}, and \textbf{1 ceiling preset}. Each preset is encoded in a JSON-based structure that specifies material assignment, tiling patterns, grout spacing, plank or tile dimensions, UV normalization, and optional micro-perturbations. An example wall preset used to create a brick surface is shown below:

Similarly, floor presets define patterns such as wood planks, ceramic tiles, marble, or carpet, while the ceiling preset controls smoothness, panel style, and material reflectance. After mesh construction, the system applies preset-driven refinements---including randomized offsets, micro-scale geometric variation, and adaptive UV remapping---followed by PBR material binding. This unified workflow yields an architecturally coherent, stylistically consistent, and procedurally controllable apartment envelope that serves as the foundation for subsequent opening construction and object-level population.

\paragraph{Room-Level Opening Construction}. Following the construction of the apartment-level envelope, we generate functional openings---including doors, windows, and other aperture types---directly from the declarative opening specifications encoded in IDSL. Each opening entry provides its type, geometric parameters, placement anchor, and orientation, which are used to instantiate both the geometric cut on the wall mesh and the parametric 3D model of the opening itself.

Given an opening specification $\mathcal{O}_j$ associated with wall segment $\mathcal{W}_i$, we apply a robust Boolean-based carving operator:
\[
\mathcal{W}_i' = \mathcal{W}_i \setminus \mathcal{O}_j,
\]
where the volume representation of $\mathcal{O}_j$ is expanded with a small safety margin to ensure stable Boolean execution. Prior to carving, the wall mesh undergoes topology cleaning to remove sliver faces and nearly coplanar triangles, preventing degeneracies in the Boolean result. For multi-room adjacency or T-junction wall configurations, a dependency-resolved ordering scheme ensures that all openings remain consistent across connected wall segments.

After the aperture is created, a parametric opening generator is invoked to synthesize the door or window asset. Our system includes a library of parameterized templates covering sliding doors, single and double hinged doors, casement windows, fixed windows, bay window structures, and high-sill variants. Each template exposes a parameter vector $\theta_{\text{open}}$ controlling frame thickness, sash geometry, sill height, panel style, subdivisions, and material configuration. The generator produces a complete manifold mesh aligned with the carved opening, ensuring that the frame and sash match the aperture geometry without gaps or overlaps.

A representative portion of an opening preset is shown below:

\begin{lstlisting}[
    language=json,
    basicstyle=\small\ttfamily,
    breaklines=true,
    style=idslstyle,
    frame=single
]
{
  "type": "single_door",
  "frame": {
    "width": 0.08,
    "depth": 0.04,
    "material": "Wood Oak"
  },
  "panel": {
    "style": "solid",
    "thickness": 0.035
  },
  "sill_height": 0.0,
  "hinge": "left",
  "swing_direction": "inward"
}
\end{lstlisting}

Once instantiated, the opening asset is automatically aligned and fitted to the carved aperture using its anchor transform and hinge orientation. PBR materials are assigned according to the preset, and UVs are adjusted to maintain consistent texel density across frames and panels. This procedure yields structurally coherent openings that integrate seamlessly with the procedurally generated wall envelope and remain fully consistent with the spatial semantics defined in IDSL.

\paragraph{Object-Level Asset Population.} At the object level, we instantiate all furniture and scene elements specified in IDSL by combining large-scale curated static assets with procedural generation modules. Each IDSL object entry provides a semantic category, a target 3D bounding box, orientation hints, a room-level style descriptor, and the final placement transform. 
% These attributes guide both our retrieval-based asset selection and the downstream placement logic.

We construct a high-quality indoor asset corpus by collecting and manually annotating a large number of static 3D assets from diverse real-world interior scenes. For each asset, we record detailed metadata including file path, asset name, semantic category, natural-language description, stylistic tag, physical dimensions (bbox$_x$, bbox$_y$, bbox$_z$), orientation flags, material attributes, modularity, articulation indicators (e.g., moving parts), typical usage scenario, free-text notes, the parent record ID, and exact bounding extents. For example, two annotated bed assets are represented as follows (translated from Chinese descriptions):

\begin{lstlisting}[
    language=json,
    basicstyle=\small\ttfamily,
    breaklines=true,
    style=idslstyle,
    style=idslstyle,
    frame=single
]
{
  "file_path": "/bed/modern_bed_001.blend",
  "asset_name": "modern_bed_001",
  "category": "bed",
  "description": "A minimalist light-gray bed featuring multiple gray pillows and a textured blanket. The color palette is soft and modern.",
  "style": "modern",
  "bbox_x": 2.6198, "bbox_y": 2.4143, "bbox_z": 1.0387,
  "orient_y_change": "yes",
  "material_type": "wood",
  "has_texture": true,
  "is_modular": false,
  "has_moving_parts": false,
  "usage_scene": "home",
  "notes": "",
  "parent_record": "",
  "min_x": -1.3000, "max_x": 1.3198,
  "min_y": -1.0289, "max_y": 1.3853,
  "min_z": 0.0060,  "max_z": 1.0447
}
{
  "file_path": "/bed/modern_bed_002.blend",
  "asset_name": "modern_bed_002",
  "category": "bed",
  "description": "A contemporary light-wood bed with an extended side-frame design and a dark-gray upholstered headboard. Soft gray bedding completes the modern look.",
  "style": "modern",
  "bbox_x": 2.6608, "bbox_y": 2.0000, "bbox_z": 0.9097,
  "orient_y_change": "yes",
  "material_type": "wood",
  "has_texture": true,
  "is_modular": "partially",
  "has_moving_parts": false,
  "usage_scene": "home",
  "notes": "",
  "parent_record": "",
  "min_x": -1.3304, "max_x": 1.3304,
  "min_y": -1.0000, "max_y": 1.0000,
  "min_z": 0.0639,  "max_z": 0.9737
}
\end{lstlisting}

Before indexing the assets, we perform standardized preprocessing, including (1) renaming assets using a canonical naming scheme, (2) reorienting the meshes to a consistent forward-facing direction, and (3) rendering a normalized front-view image for visual feature extraction. BLIP-2 embeddings computed from these renders provide category-independent semantic representations that significantly improve retrieval robustness. Assets sourced from external repositories such as Objaverse undergo the same processing pipeline, ensuring consistent metadata, front-view rendering, and embedding extraction.

At retrieval time, given an IDSL object $o_i$, candidate assets are ranked using a multi-modal matching function combining semantic similarity, geometric compatibility, and stylistic coherence:
\[
\begin{aligned}
\text{score}(a, o_i) = \ &
\lambda_{\text{sem}} \cdot \text{sim}_{\text{BLIP}}(f_a, f_{o_i}) \\
& + \lambda_{\text{geo}} \cdot \text{IoU}(\text{obb}_a, \text{obb}_{o_i}) \\
& + \lambda_{\text{style}} \cdot \psi(s_a, s_{\text{room}}).
\end{aligned}
\]

If the top-ranked asset falls below a threshold $\tau$, the system invokes a corresponding procedural generator from our library of 27 category-specific PCG modules. These generators synthesize geometry consistent with the target bounding box and stylistic parameters, enabling flexible and reliable instantiation even when no suitable asset is found in the static corpus.

Once selected or generated, the asset is placed in the scene using the transform specified in IDSL. The system then performs clearance checking against structural elements and nearby objects, resolves orientation according to the standardized forward direction, assigns PBR materials, and normalizes UV coordinates to maintain consistent texel density across the scene. This hybrid retrieval-and-generation pipeline ensures that all object-level specifications in IDSL are matched with high-fidelity, contextually compatible assets, supporting both large-scale diversity and precise controllability.

\subsection{Relationship-Aware Post-Placement Optimization}

Once all objects are instantiated in the scene according to their IDSL-specified transforms, we perform a relationship-aware post-processing step to correct positional inaccuracies and enforce physically plausible spatial relationships. Because retrieved or procedurally generated assets may not perfectly match the target bounding box or proportions expected by IDSL, small geometric discrepancies can accumulate. These deviations may violate semantic relations—for example, cups floating above a table surface if the retrieved table is lower than expected or if its top surface does not align with the predicted support plane. To address this, each object is associated with a set of relational constraints extracted from IDSL in the form:

\begin{lstlisting}[
    language=json,
    basicstyle=\small\ttfamily,
    breaklines=true,
    style=idslstyle,
    frame=single
]
"relations": [
  {
    "relation_graph": {
      "child_tags": ["-Subpart(top)", "Subpart(back)", "-Subpart(front)"],
      "parent_tags": ["Subpart(wall)", "-Subpart(support)",
                      "Subpart(visible)", "-Subpart(ceiling)"],
      "margin": 0,
      "check_z": true,
      "rev_normal": false,
      "relation_type": "StableAgainst"
    },
    "target_name": "kitchen_0/0",
    "child_plane_idx": 0,
    "parent_plane_idx": 0,
    "value": null
  }
]
\end{lstlisting}

Each relation defines a pair of interacting surfaces (child and parent), their semantic subpart labels (e.g., support, top, back, visible), and the constraint type such as \texttt{StableAgainst}, \texttt{OnTopOf}, \texttt{AlignedWith}, or \texttt{AdjacentTo}. For a given object $o_i$, we retrieve all other objects in the scene that satisfy these relational descriptors and identify the corresponding geometric planes. Let $P_i^{\text{child}}$ and $P_j^{\text{parent}}$ denote the matched planes from object $o_i$ and its relational partner $o_j$. The optimization adjusts the placement transform of $o_i$ to minimize spatial violations:

\[
\begin{aligned}
\min_{\mathbf{p}_i, \mathbf{q}_i} \ \ 
& d\!\left(T_i(P_i^{\text{child}}),\; T_j(P_j^{\text{parent}})\right) \\
& + \lambda_{\text{stab}} \, E_{\text{stability}}(o_i) \\
& + \lambda_{\text{coll}} \, E_{\text{collision}}(o_i).
\end{aligned}
\]

where the first term enforces geometric alignment between relational surfaces (e.g., cup bottom plane aligning with table top plane), $E_{\text{stability}}$ ensures the object sits naturally under gravity (eliminating floating or sinking artifacts), and $E_{\text{collision}}$ penalizes interpenetration with neighboring geometry. Optimizations are solved via projected gradient descent with small-step updates to preserve placement intent while correcting errors from asset mismatch.

\begin{figure}[t]
\centering
\includegraphics[width=\linewidth]{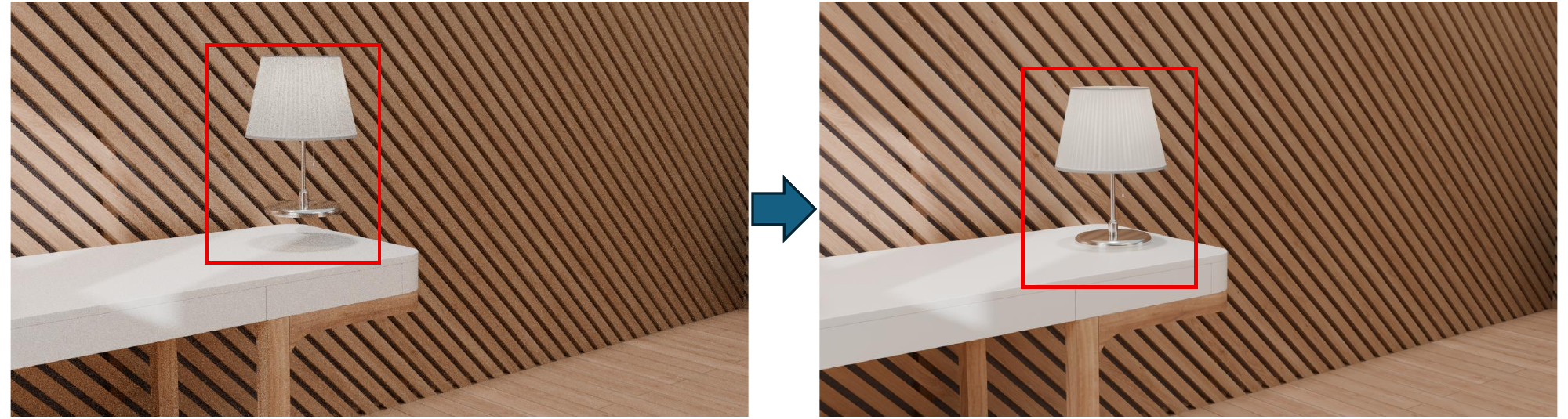}
\caption{
Example of relationship-aware post-placement optimization. 
Left: before optimization, the desk lamp floats above the tabletop due to geometric mismatch between the retrieved asset and the IDSL-specified support relation. 
Right: after optimization, the lamp is precisely aligned and stably placed on the table surface, 
demonstrating the correction of support relations and elimination of floating artifacts.
}
\label{fig:post_optimization_example}
\end{figure}

This process yields significant improvements in physical plausibility and scene realism. 
Support relations are correctly enforced (objects rest firmly on supporting surfaces), 
and adjacency or alignment relations are respected 
(e.g., chairs slide under desks, appliances snap to counters). 
Floating, sinking, or intersecting artifacts are effectively eliminated. 
Figure~\ref{fig:post_optimization_example} shows a representative example: 
before optimization, a desk lamp matched from the asset corpus floats above the tabletop due to geometric mismatch; 
after applying our relationship-aware correction, the lamp is accurately aligned to the table’s top surface, 
demonstrating the effectiveness of our post-processing strategy. 
As a result, the final scene adheres not only to the semantic intent of IDSL 
but also to practical geometric coherence, enabling downstream rendering, simulation, and VR/AR applications 
to operate on structurally valid environments.

% \subsection{Supplementary Explanation}
% This section provides explanation of the symbols in Section \ref{sec:hierarchical indoor generation}.

\section{Details of Experiments}
\label{sec:details of experiments_supp}

\subsection{Controllability Metrics}

To quantitatively evaluate controllable scene generation, we employ three metrics computed from the unified IDSL-based evaluation pipeline: 
Layout Fidelity (LF), 
Constraint Satisfaction Rate on object count (CSR\textsubscript{count}), 
and Constraint Satisfaction Rate on relational constraints (CSR\textsubscript{rel}).

Layout Fidelity (LF).
LF measures how faithfully a generated layout follows a target spatial configuration.
We compute LF by converting both the reference and generated scenes into 
semantic masks and computing a category-wise Intersection-over-Union (IoU).

Data preparation.
From each IDSL scene, we extract object categories, 3D bounding boxes 
(projected onto the floor plane), and the room polygon to define a unified 
spatial domain for rasterization.

Semantic rasterization.
Both the reference and generated layouts are rasterized onto a fixed-resolution grid.
For each object, its 2D footprint 
$[x_{\min}, x_{\max}] \times [y_{\min}, y_{\max}]$ 
is filled with the corresponding semantic category ID, producing 
$M_{\text{ref}}$ and $M_{\text{gen}}$.

\textbf{IoU computation.}
For each category $c$, we compute:
\begin{equation}
\text{IoU}_c =
\frac{|M_{\text{ref}} = c \wedge M_{\text{gen}} = c|}
     {|M_{\text{ref}} = c \vee M_{\text{gen}} = c|}.
\label{eq:iou_c}
\end{equation}

LF is the mean over all categories:
\begin{equation}
\text{LF} = 
\frac{1}{|\mathcal{C}|} 
\sum_{c \in \mathcal{C}} \text{IoU}_c.
\label{eq:lf}
\end{equation}

This metric captures spatial controllability by penalizing misplaced, missing, 
or geometrically inconsistent objects, and complements CSR\textsubscript{count} 
and CSR\textsubscript{rel} which evaluate constraint satisfaction.

CSR\textsubscript{count}.
CSR\textsubscript{count} evaluates whether the generated scene satisfies the object-count constraints.
Let $n_{c}^{\text{ref}}$ and $n_{c}^{\text{gen}}$ denote reference and generated counts.
We compute the strict version as:
\begin{equation}
\text{CSR}_{\text{count}}^{\text{strict}}
= \frac{1}{|\mathcal{C}|}
\sum_{c \in \mathcal{C}}
\mathbb{1}\left[n_{c}^{\text{ref}} = n_{c}^{\text{gen}}\right].
\label{eq:csr_count_strict}
\end{equation}

The soft version is:
\begin{equation}
\text{CSR}_{\text{count}}^{\text{soft}}
= \frac{1}{|\mathcal{C}|}
\sum_{c \in \mathcal{C}}
\frac{\min(n_{c}^{\text{ref}}, n_{c}^{\text{gen}})}
     {\max(n_{c}^{\text{ref}}, n_{c}^{\text{gen}})}.
\label{eq:csr_count_soft}
\end{equation}

CSR\textsubscript{rel}.
CSR\textsubscript{rel} measures whether geometric and functional constraints are satisfied, including Facing, Near, and StableAgainst.
It is computed as:
\begin{equation}
\text{CSR}_{\text{rel}} 
= \frac{1}{|\mathcal{R}|}
\sum_{r \in \mathcal{R}}
\mathbb{1}\left[\text{constraint}(r)=\text{satisfied}\right].
\label{eq:csr_rel}
\end{equation}

Each relational constraint is checked using the following rules:
\begin{equation}
\Delta\theta \le \theta_{\max} 
\quad (\text{Facing}),
\label{eq:facing}
\end{equation}

\begin{equation}
\lVert p_a - p_b \rVert_2 \le d_{\max} 
\quad (\text{Near}),
\label{eq:near}
\end{equation}

\begin{equation}
\text{dist}(p, \partial\Omega) \le m_{\max} 
\quad (\text{StableAgainst}).
\label{eq:stableagainst}
\end{equation}

These three metrics jointly capture the spatial, semantic, and relational controllability of the generated scenes.

\subsection{CAD Geometric Fidelity Metrics}
\label{appendix:cad_metrics}

\noindent Wall Accuracy (Wall Acc.) measures the 
geometric alignment between generated wall polygons and 
the input CAD floor plan, computed as the Intersection 
over Union (IoU) between the generated and reference wall 
regions.

\noindent Room Area Error (Area Err.) measures 
the relative area deviation of each generated room from 
its CAD annotation, computed as:
\begin{equation}
\text{Area Err.} = \frac{1}{N}\sum_{i=1}^{N} 
\frac{|A_i^{gen} - A_i^{gt}|}{A_i^{gt}}
\end{equation}
where $A_i^{gen}$ and $A_i^{gt}$ denote the generated 
and ground-truth area of the $i$-th room, and $N$ is 
the total number of rooms.

\noindent \textbf{Opening Accuracy (Open. Acc.).}
Opening Accuracy measures how accurately generated doors and windows are placed with respect to the CAD annotations. 
For each opening, we compare the center position of the generated opening with that of the corresponding ground-truth opening. 
An opening is considered correctly placed if the Euclidean distance between the two positions is smaller than a predefined threshold $\delta$. 
The metric is defined as
\begin{equation}
\text{Open. Acc.} = \frac{1}{M}\sum_{j=1}^{M}
\mathbf{1}\!\left[\left\|p_j^{gen} - p_j^{gt}\right\| < \delta\right],
\end{equation}
where $p_j^{gen}$ and $p_j^{gt}$ denote the generated and ground-truth center positions of the $j$-th opening, respectively, and $M$ is the total number of openings. 
Here, $\mathbf{1}[\cdot]$ is an indicator function that equals 1 if the condition is satisfied and 0 otherwise. 
A higher Open. Acc. indicates that a larger proportion of openings are placed close to their correct positions.

\section{Additional Experimental Results}
\label{sec:additional experimental results_supp}

\subsection{Quantitative Evaluation with SceneEval}
\label{sec:Quantitative Evaluation with SceneEval}

\begin{table*}[t]
\centering
\caption{
\textbf{Evaluation with text-image scores and SceneEval metrics}~\cite{tam2026sceneeval}.
Overall, RoomPilot outperforms prior work on metrics measuring the fidelity of object placements relative to the input text, including Text-Image Score and SceneEval Fidelity metrics. 
For plausibility metrics, RoomPilot achieves the highest support rate while maintaining strong physical validity and scene completeness. 
Although LayoutGPT and InstructScene obtain lower collision rates or slightly better navigation, they place substantially fewer objects per scene, resulting in less complete layouts.
Bold indicates highest results.
}
\resizebox{\linewidth}{!}
{
\begin{tabular}{@{} l rrr rrrr rrrrrr r @{}}
\toprule

& \multicolumn{3}{c}{Text-Image Score} 
& \multicolumn{4}{c}{SceneEval Fidelity} 
& \multicolumn{6}{c}{SceneEval Plausibility} 
& \multirow{2.5}{*}{\shortstack{Avg.\ $\#$Obj\\ per Scene}}
\\ 
  
\cmidrule(l){2-4} \cmidrule(l){5-8} \cmidrule(lr){9-14} 
& $\uparrow~$BLIP & $\uparrow~$CLIP  & $\uparrow~$VQA

& $\uparrow~$CNT$_{\%}$ & $\uparrow~$ATR$_{\%}$ & $\uparrow~$OOR$_{\%}$ & $\uparrow~$OAR$_{\%}$

& $\downarrow~$COL$_{ob\%}$ & $\downarrow~$COL$_{sc\%}$ & $\uparrow~$SUP$_{\%}$ & $\uparrow~$NAV$_{\%}$
& $\uparrow~$ACC$_{\%}$ & $\downarrow~$OOB$_{\%}$

\\
\midrule

LayoutGPT
& 0.0613 & 0.1670 & 0.2964
& 19.54 & 18.98 &  2.87 &   5.24
& \textbf{12.96} & \textbf{30.00} & 28.24 & \textbf{100.00}
& 47.29 & 73.11 
& 5.17
\\

InstructScene                           
& 0.0845 & 0.1681 & 0.4082
& 25.48 & 22.26 & 11.17 & 10.48
& 51.18 & 84.00 & 75.09 & 99.53
& 77.30 & 22.92
& 8.07
\\
\midrule

LayoutVLM
& 0.0857 & 0.1612 & 0.3268
& 41.19 & 22.26 & 8.60 & 23.29
& 36.09 & 69.00 & 67.96 & 98.75
& 85.91 & 4.14
& 11.36
\\

Holodeck
& 0.1230 & 0.1820 & 0.5549
& 44.64 & 39.42 & 20.92 & 49.60
& 17.32 & 73.00 & 62.12 & 99.45
& \textbf{90.55} & \textbf{1.30}
& 24.71
\\

HSM
& 0.1748 & 0.1841 & 0.5627
& 61.30 & 59.49 & 40.40 & 70.28
& 16.42 & 61.00 & 85.44 & 98.97
& 86.80 & 2.13
& 20.65
\\

RoomPilot (ours)
& \textbf{0.1816} & \textbf{0.1863} & \textbf{0.5714}
& \textbf{64.85} & \textbf{61.72} & \textbf{43.18} & \textbf{73.94}
& 15.87 & 38.00 & \textbf{88.62} & 99.21
& 89.34 & 1.76
& 22.48
\\

\bottomrule
\end{tabular}
}
\label{tab:scene_quantitative}
\end{table*}

As shown in Table~\ref{tab:scene_quantitative}, following the evaluation protocol of SceneEval~\cite{tam2026sceneeval}, RoomPilot delivers the strongest overall performance on text-image alignment and scene fidelity metrics. It achieves the best results on BLIP, CLIP, and VQA, suggesting better consistency between the generated scenes and the input descriptions. RoomPilot also performs best on CNT, ATR, OOR, and OAR, indicating more accurate object composition and stronger adherence to textual constraints.
For plausibility metrics defined in SceneEval~\cite{tam2026sceneeval}, RoomPilot attains the highest support rate and remains competitive on collision, navigation, accuracy, and out-of-boundary measures. While some baselines achieve slightly better results on individual plausibility metrics, they often generate fewer objects per scene, resulting in less complete layouts. By contrast, RoomPilot preserves strong physical plausibility while producing richer and more complete scenes, demonstrating a favorable trade-off between fidelity and scene completeness.

\subsection{LLM Parameter Parsing Evaluation}

\begin{table*}[t]
\centering
\caption{
LLM-based IDSL parsing across three prompt settings and two parsing methods.
LLM-based IDSL parsing performance across three prompt settings and two parsing methods. 
Semantic Accuracy evaluates correctness over room types (Room Type), object categories (Obj Class), and spatial relations (Rel Match). 
Structural Proximity measures object-count fidelity (Obj Count) and spatial layout deviation (Layout Error). 
Topological Consistency assesses room-adjacency correctness (Room Adj.) and door/window-to-room assignment (Open Assign).
}
\resizebox{\textwidth}{!}{
\begin{tabular}{llccccccc}
\toprule
\multicolumn{2}{l}{Setting / Method} &
\multicolumn{3}{c}{Semantic Accuracy$\uparrow$} &
\multicolumn{2}{c}{Structural Proximity$\uparrow$} &
\multicolumn{2}{c}{Topological Consistency$\uparrow$} \\
\cmidrule(lr){3-5}
\cmidrule(lr){6-7}
\cmidrule(lr){8-9}
& &
Room Type 
& Obj Class 
& Rel Match 
& Obj Count 
& Layout Error 
& Room Adj. 
& Open Assign \\
\midrule

\multirow{2}{*}{Simple Single-Room} 

& CodeOnly 
& 0.74 & 0.65 & 0.62
& 0.54 & 0.42 
& --    & 0.57 \\
& LLM (Ours) 
& \textbf{0.93} & \textbf{0.87} & \textbf{0.85}
& \textbf{0.75} & \textbf{0.62} 
& \textbf{--}    & \textbf{0.86} \\
\midrule
\multirow{2}{*}{Complex Single-Room} 

& CodeOnly 
& 0.65 & 0.63 & 0.55
& 0.61 & 0.35 
& --    & 0.46 \\
& LLM (Ours) 
& \textbf{0.89} & \textbf{0.82} & \textbf{0.81}
& \textbf{0.79} & \textbf{0.69} 
& \textbf{--}    & \textbf{0.57} \\
\midrule

\multirow{2}{*}{Multi-Room Suite} 

& CodeOnly 
& 0.69 & 0.61 & 0.58
& 0.59 & 0.49 
& 0.64 & 0.50 \\
& LLM (Ours) 
& \textbf{0.85} & \textbf{0.80} & \textbf{0.78}
& \textbf{0.78} & \textbf{0.79} 
& \textbf{0.87} & \textbf{0.90} \\
\bottomrule
\end{tabular}
}
\label{tab:llm_idsl_parsing_ablation}
\end{table*}

As shown in Table~\ref{tab:llm_idsl_parsing_ablation}, we conducted the evaluation using 50 test cases across all three levels, employing the GPT-4o model to assess the parsing capability of the LLM component. 
Table~\ref{tab:llm_idsl_parsing_ablation} evaluates LLM-based IDSL parsing across seven metrics grouped into three categories. 
\textbf{Semantic Accuracy} measures the correctness of predicted room types (\textit{Room Type}), object categories (\textit{Obj Class}), and spatial relations (\textit{Rel Match}). 
\textbf{Structural Proximity} quantifies the structural fidelity between predictions and ground truth, including object-count correctness (\textit{Obj Count}) and spatial layout deviation based on object-center distances (\textit{Layout Error}). 
\textbf{Topological Consistency} evaluates higher-level scene-graph correctness: room adjacency (\textit{Room Adj.}) in multi-room configurations and door/window-to-room assignment accuracy (\textit{Open Assign}).

The LLM-based parser consistently outperforms the rule-based \textit{CodeOnly} baseline across all three prompt settings, demonstrating clear improvements in nearly every metric.  
The enhancements in \textbf{Semantic Accuracy} reflect that the LLM more reliably captures room semantics, object categories, and spatial relations, particularly as descriptions become more detailed.  
In \textbf{Structural Proximity}, the LLM consistently performs better in object-count fidelity and layout error, suggesting that it is more capable of inferring object presence and approximating their spatial arrangement directly from natural language input.  
For \textbf{Topological Consistency}, the LLM achieves higher accuracy in predicting room adjacency and assigning door/window openings to the correct room boundaries, especially in more complex multi-room configurations.  
% Overall, these results confirm that incorporating LLM reasoning significantly enhances the quality of IDSL parsing, resulting in more accurate, coherent, and topologically consistent scene representations compared to the rule-based method alone.

\begin{figure}[t]
    \centering
    \includegraphics[width=\columnwidth]{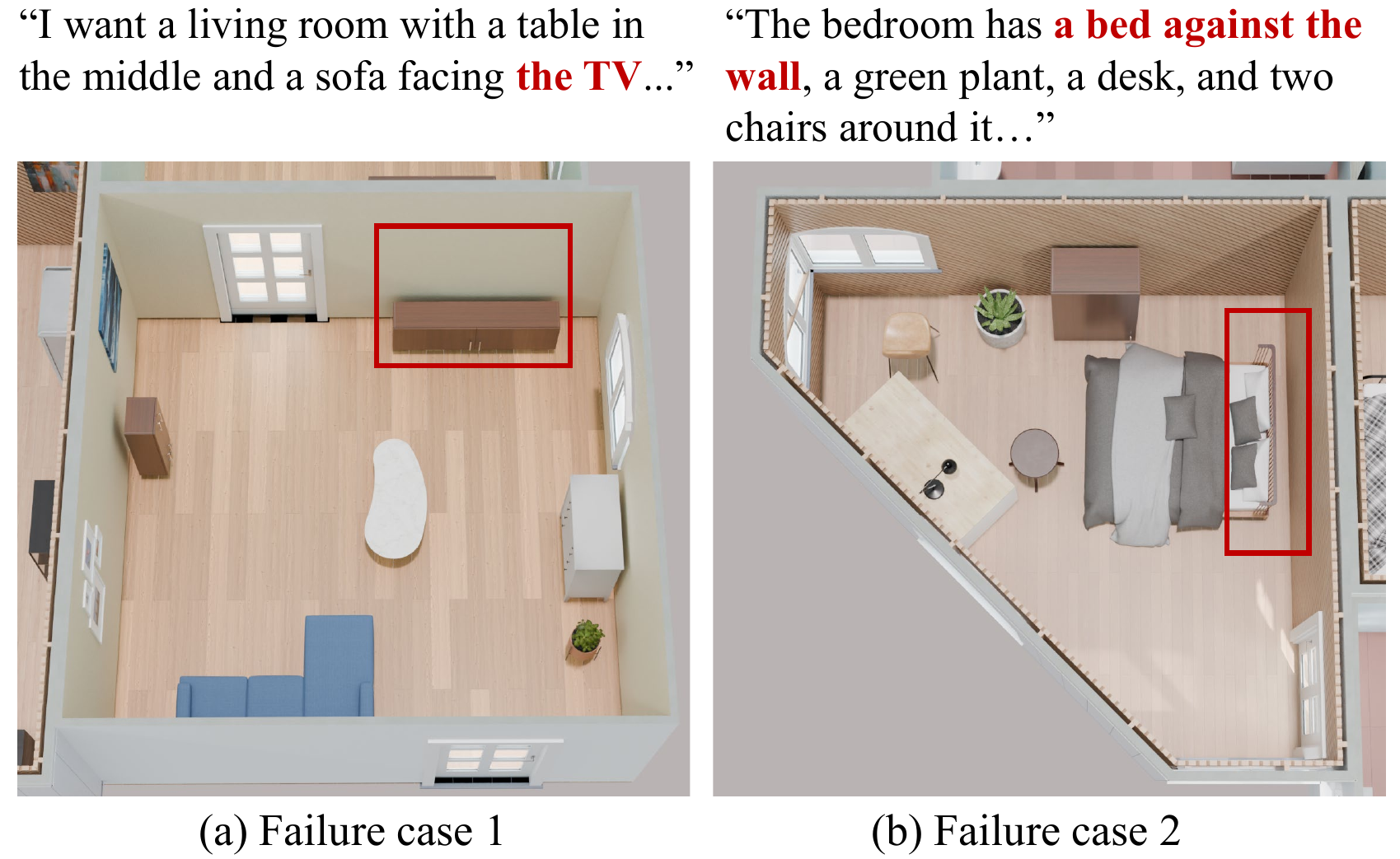}
    \caption{
    Text-conditioned grounding failure cases.
    (a) ``sofa facing the TV'' is specified, but no TV is instantiated.
    (b) ``bed against the wall'' is specified, but the bed remains offset from the wall.
    }
    \label{fig:failure_cases}
\end{figure}

\subsection{Runtime Analysis}
\label{sec:runtime}

\begin{table}[t]
\centering
\caption{Runtime analysis of RoomPilot. We report the average runtime in seconds and the average number of placed objects per scene.}
\resizebox{0.8\linewidth}{!}{
\begin{tabular}{lcc}
\toprule
Method & Runtime $\downarrow$ & \#Objects $\uparrow$ \\
\midrule
RoomPilot (Text) & 2968.42 & 31.48 \\
RoomPilot (CAD) & 2714.35 & 30.76 \\
w/o Optimization & 2441.67 & 28.93 \\
w/o Semantic Energy & 2826.51 & 30.85 \\
w/o Structural Energy & 2798.44 & 31.02 \\
w/o Adaptive Annealing & 2879.63 & 31.11 \\
w/o Prog.\ Rule Activation & 2896.20 & 30.97 \\
\bottomrule
\end{tabular}}
\label{tab:runtime}
\end{table}

We analyze the runtime of RoomPilot across different configurations. Since RoomPilot is built on top of a procedural indoor synthesis backend while additionally introducing multimodal semantic parsing and self-regulating layout optimization, its runtime is moderately higher than that of a pure procedural pipeline. In particular, text-conditioned generation is slower than CAD-conditioned generation because it requires one building-level parsing pass and multiple room-level LLM parsing calls, whereas CAD input directly provides room geometry and only requires lightweight geometric parsing and object-level completion.
As shown in Table~\ref{tab:runtime}, the full RoomPilot system remains practical while producing complete indoor scenes with a relatively large number of placed objects. Removing the optimization stage reduces runtime noticeably, but also decreases scene completeness. Ablating individual optimization components yields smaller runtime reductions, suggesting that the added computation is justified by improved controllability and physically coherent scene synthesis.

\subsection{Failure cases}

Figure~\ref{fig:failure_cases} reports failure cases that are not able to align with textual descriptions.
These cases are attributable to the intrinsic behavior of the simulated annealing (SA) procedure adopted in the self-regulating scene optimization.
In Fig.~\ref{fig:failure_cases} (a), the textual instruction ``sofa facing the TV'' requires instantiating an additional object and introducing a new directional relation.
During optimization, SA rapidly converges to a structurally stable, collision-free furniture layout.
As the temperature decreases with increasing rule satisfaction, proposals that introduce a new object or significantly perturb the stabilized configuration are unlikely to be accepted, resulting in a final scene without a TV.
In Fig.~\ref{fig:failure_cases} (b), the constraint ``bed against the wall'' competes with accessibility and collision-related structural energies in an irregular room geometry.
SA resolves this competition by settling at a near-wall configuration early in the search; enforcing strict wall contact would temporarily increase structural energy and is therefore rejected after cooling.
These cases arise from SA’s characteristic freeze-out behavior when jointly optimizing discrete object composition and continuous spatial relations under competing energy terms. We will further improve it in our future work.

\subsection{Additional Visualizations}

\begin{figure*}[hbtp]
    \centering
    % \vspace{-0.8em}
    \includegraphics[width=1\linewidth]{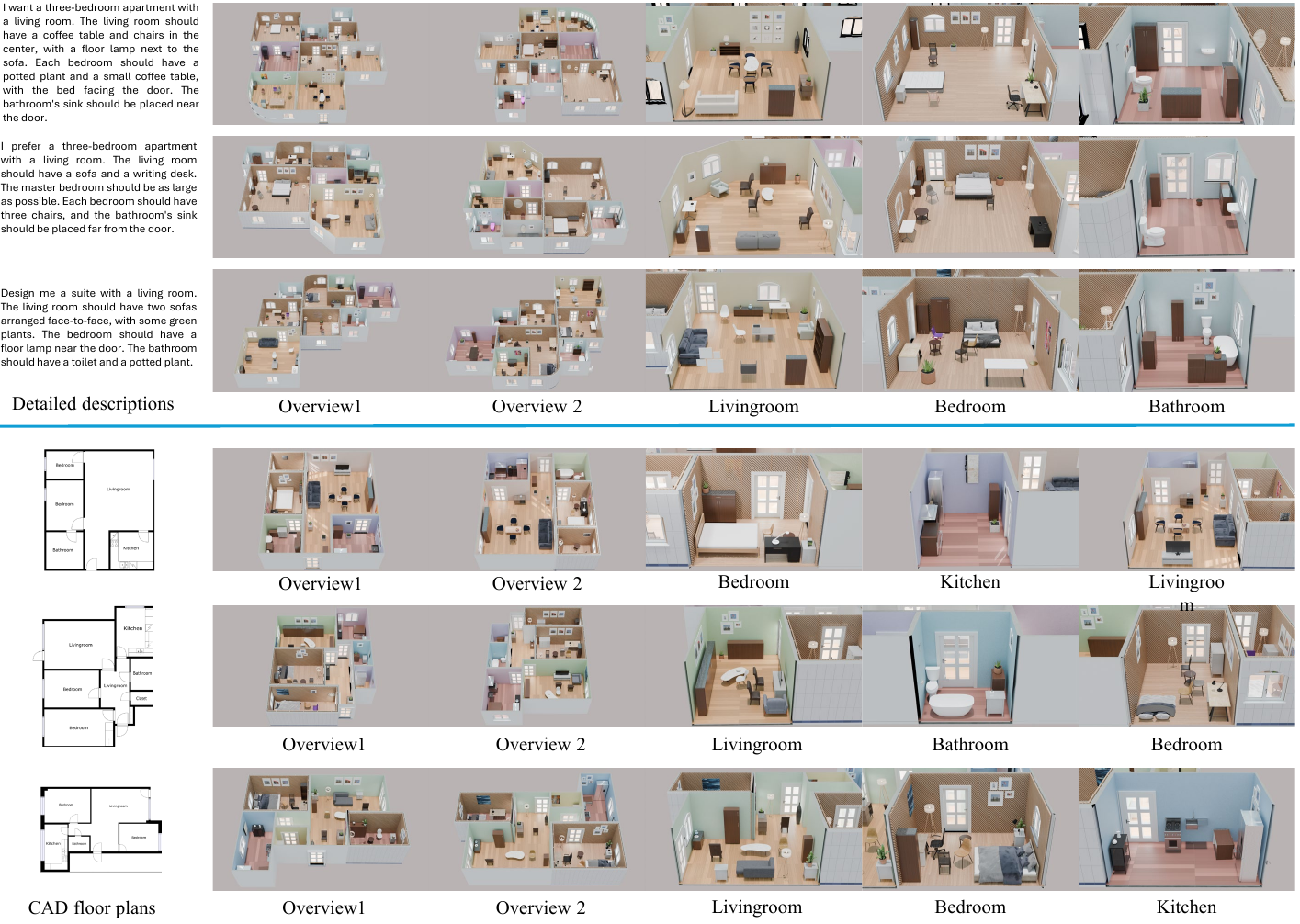}
    \caption{
Visual results of RoomPilot across different settings. The top row displays the detailed descriptions and corresponding 3D scenes. The bottom row shows the CAD floor plans for each layout, with the corresponding 3D visualizations different room types.}
    \label{fig:supple_1}
\end{figure*}

To validate the diverse design outcomes generated by RoomPilot based on varying user inputs, we present additional visual results shown in Figure \ref{fig:supple_1}. The top row features detailed descriptions of three apartment layouts, paired with their corresponding 3D visualizations. These scenes, created according to user specifications, represent various room types such as living rooms, bedrooms, bathrooms, and kitchens, demonstrating RoomPilot’s ability to transform complex design instructions into realistic spatial arrangements.
The bottom row displays the CAD floor plans for each layout, offering technical blueprints that serve as a reference for the 3D visualizations. This allows for a clear comparison between the two representations.

% \subsection{Part One}

% Lorem ipsum dolor sit amet, consectetur adipiscing elit. Morbi
% malesuada, quam in pulvinar varius, metus nunc fermentum urna, id
% sollicitudin purus odio sit amet enim. Aliquam ullamcorper eu ipsum
% vel mollis. Curabitur quis dictum nisl. Phasellus vel semper risus, et
% lacinia dolor. Integer ultricies commodo sem nec semper.

% \subsection{Part Two}

% Etiam commodo feugiat nisl pulvinar pellentesque. Etiam auctor sodales
% ligula, non varius nibh pulvinar semper. Suspendisse nec lectus non
% ipsum convallis congue hendrerit vitae sapien. Donec at laoreet
% eros. Vivamus non purus placerat, scelerisque diam eu, cursus
% ante. Etiam aliquam tortor auctor efficitur mattis.

% \section{Online Resources}

% Nam id fermentum dui. Suspendisse sagittis tortor a nulla mollis, in
% pulvinar ex pretium. Sed interdum orci quis metus euismod, et sagittis
% enim maximus. Vestibulum gravida massa ut felis suscipit
% congue. Quisque mattis elit a risus ultrices commodo venenatis eget
% dui. Etiam sagittis eleifend elementum.

% Nam interdum magna at lectus dignissim, ac dignissim lorem
% rhoncus. Maecenas eu arcu ac neque placerat aliquam. Nunc pulvinar
% massa et mattis lacinia.

\end{document}